\documentclass[journal]{IEEEtran}
\usepackage[margin=1in]{geometry}
\usepackage{cite}
\usepackage{float}
\usepackage{amsmath}
\usepackage{amsthm}
\usepackage{amssymb}
\usepackage{graphicx,wrapfig}
\makeatletter

\floatstyle{ruled}
\newfloat{algorithm}{tbp}{loa}
\providecommand{\algorithmname}{Algorithm}
\floatname{algorithm}{\protect\algorithmname}

\theoremstyle{plain}

\newtheorem{prop}{\protect\propositionname}
\newtheorem{thm}{\protect\theoremname}
\newtheorem{lemma}{\protect\lemmaname}

\@ifundefined{date}{}{\date{}}

\usepackage{caption}
\usepackage{dsfont,subcaption}

\newcommand{\ignore}[1]{}

\usepackage{semantic}
\mathlig{==}{\equiv}
\mathlig{=.}{\doteq}
\mathlig{:=}{\triangleq}
\mathlig{<<}{\ll}
\mathlig{>>}{\gg}
\mathlig{<>}{\neq}
\mathlig{<=}{\leq}
\mathlig{>=}{\geq}
\mathlig{<==}{\Leftarrow}
\mathlig{==>}{\Rightarrow}
\mathlig{<=>}{\Leftrightarrow}
\mathlig{<==>}{\iff}
\mathlig{<--}{\leftarrow}
\mathlig{-->}{\rightarrow}
\mathlig{<->}{\leftrightarrow}
\mathlig{+-}{\pm}
\mathlig{-+}{\mp}
\mathlig{...}{\dots}
\mathlig{!=}{\stackrel{!}{=}}

\DeclareMathOperator*{\argmin}{\arg\!\min}
\newif\ifshowanswer    % by default set to false.
\newcommand{\isitthree}[1]
{
  \ifnum#1=3
    number #1 is 3
  \else
    number #1 is not 3
  \fi
}

\newcommand{\be}{\begin{equation}}
\newcommand{\ee}{\end{equation}}

\newcommand\R{{\mathbb{R}}}

\renewcommand\P{{\mathds{P}}}
\newcommand\E{{\mathds{E}}}

\newcommand\CE{{\mathcal E}}

\newcommand\CH{{\mathcal H}}

\newcommand\CN{{\mathcal N}}

\usepackage[dvipsnames]{xcolor}

\usepackage{amsmath}
\newcommand\numberthis{\addtocounter{equation}{1}\tag{\theequation}}
\allowdisplaybreaks

\providecommand{\definitionname}{Definition}
\providecommand{\lemmaname}{Lemma}
\providecommand{\propositionname}{Proposition}
\providecommand{\theoremname}{Theorem}

\usepackage{dsfont,xcolor}
\usepackage{algorithm}
\usepackage{algpseudocode}

\usepackage{hyperref}
\newtheorem{cor}{Corollary}

\newcommand{\printfnsymbol}[1]{
  \hspace{-.15cm}\textsuperscript{\@fnsymbol{#1}}\hspace{-.05cm}
}

\newcommand{\sg}[1]{\|#1 \|_{\Psi_2}}

\begin{document}
\title{Bandit-Based Monte Carlo Optimization\\ for Nearest Neighbors}
\author{
\IEEEauthorblockN{Vivek Bagaria\printfnsymbol{1}, Tavor Z. Baharav\printfnsymbol{1}, Govinda M. Kamath\printfnsymbol{1}, David N. Tse}

\IEEEauthorblockA{Department of Electrical Engineering, Stanford University\\
\{vbagaria, tavorb, gkamath, dntse\}@stanford.edu}
}

\newlength{\textfloatsepsave}
\setlength{\textfloatsepsave}{\textfloatsep}
\setlength{\textfloatsep}{.1cm}

\maketitle

\begingroup
{\begingroup{\renewcommand{\thefootnote}{\printfnsymbol{1}}
\footnotetext{Equal contribution, listed alphabetically. Contact: Tavor Z. Baharav.\\ The authors gratefully acknowledge funding from the NSF GRFP, Alcatel-Lucent Stanford Graduate Fellowship, NSF grant under CCF-1563098, and the Center for Science of Information (CSoI), an NSF Science and Technology Center under grant agreement CCF-0939370.}
}\endgroup}
\begin{abstract}

The celebrated Monte Carlo method estimates an expensive-to-compute quantity by random sampling. Bandit-based Monte Carlo optimization is a general technique for computing the minimum of many such expensive-to-compute quantities by adaptive random sampling. The technique converts an optimization problem into a statistical estimation problem which is then solved via multi-armed bandits. 
We apply this technique to solve the problem of high-dimensional $k$-nearest neighbors, developing an algorithm which we prove is able to identify exact nearest neighbors with high probability. 
We show that under regularity assumptions on a dataset of $n$ points in $d$-dimensional space, the complexity of our algorithm scales logarithmically with the dimension of the data as $O\left((n+d)\log^2 \left(\frac{nd}{\delta}\right)\right)$ for error probability $\delta$, rather than linearly as in exact computation requiring $O(nd)$.
We corroborate our theoretical results with numerical simulations, showing that our algorithm outperforms both exact computation and state-of-the-art algorithms such as kGraph, NGT, and LSH on real datasets.
\end{abstract}

\section{Introduction}
The use of random sampling to convert the problem of {\em computing} a deterministic quantity to efficiently {\em estimating} it dates back to the Buffon's needle experiment of Mario Lazzarini in 1901, and was later developed into the celebrated {\em  Monte Carlo method} by Stanislav Ulam. 
A direct application of the method to solve an optimization problem of the form:
\begin{equation}\label{eq:prob_formulation_intro}
    \underset{i \in \mathcal{I}}{\text{argmin}} \ \theta_i
\end{equation}
is to generate enough samples (Figure \ref{fig:block_dg}(a)) to estimate each $\theta_i$ accurately and then compute the minimum of the estimates (Figure \ref{fig:block_dg}(b)).
However, this is computationally inefficient if the set $\mathcal{I}$ is large, since the $\theta_i$'s which are much greater than the minimum need not be estimated as accurately as the $\theta_i$'s which are closer to the minimum.
Instead, a more efficient procedure is to first estimate all the $\theta_i$'s crudely using few samples and then {\em adaptively} focus the sampling on the more promising candidates (Figure \ref{fig:block_dg}(c)).
Efficient algorithms to perform this adaptive sampling can be obtained by reformulating this into a {\em multi-armed bandit} problem, where each arm corresponds to an element in the set $\mathcal{I}$, each arm's mean corresponds to $\theta_i$, and each pull of arm $i$ corresponds to generating a Monte Carlo sample of $\theta_i$.

\begin{figure}[h]
  \begin{center}
  \vspace{-.4cm}
    \includegraphics[width=\linewidth]{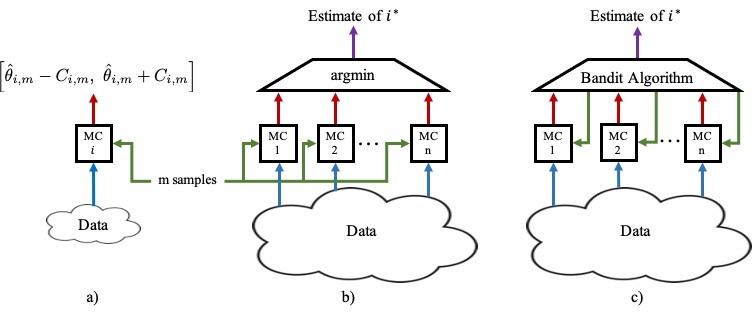}
  \vspace{-.7cm}
  \caption{(a) A \textit{Monte Carlo
  box} (MC $i$) gives an estimate $\hat{\theta}_{i,m}$ of $\theta_i$ together with a confidence interval, based on $m$ samples from the data. (b) A non-adaptive Monte Carlo optimization technique which uses a fixed number of samples to uniformly estimate the $\theta_i$'s. The Monte Carlo box is invoked once for each $i$.  (c) A bandit-based Monte Carlo optimization technique adaptively invokes the Monte Carlo boxes to estimate the $\theta_i$'s to the necessary accuracy.
}\label{fig:block_dg}
\end{center}
\vspace{-.3cm}
\end{figure}

Instances of this general technique of {\em bandit-based Monte Carlo optimization } (BMO) have already appeared in the machine learning literature.
One example is Monte Carlo Tree Search, proposed by \cite{OR_folks,KocSze} as a method to solve large-scale Markov Decision 
Problems. Here $\mathcal{I}$ is the set of all possible actions at a state $s$ and $\theta_i = Q(i,s)$, the expected value of the total reward of taking action $i$ at state $s$ and then following the optimal policy.
Another example that follows the principles of BMO is adaptive hyper-parameter tuning for neural networks \cite{JamTal,LiJamEtAl}. Here the set $\mathcal{I}$ consists of possible hyper-parameter configurations and $\theta_i$ is the validation performance of the neural network under the $i$-th hyper-parameter configuration. %, where pulling arm $i$ corresponds to training the $i$-th neural network for 1 unit of time, improving its validation performance. 
A third example is computing the medoid of $n$ points in high-dimensional space\cite{BagKamNtrZhaTse,baharav2019Medoid}.
Here, the set $\mathcal{I}$ consists of all points in the dataset and $\theta_i$ is the average distance from point $i$ to all other points.
In this application, the Monte Carlo box (Figure \ref{fig:block_dg}(a)) for estimating $\theta_i$ randomly samples $m$ points and computes the average distance $\hat{\theta}_{i,m}$ from point $i$ to these $m$ points. This estimator is unbiased and can be efficiently updated when more sampling is needed.

In this paper we demonstrate the power and broad applicability of the BMO technique by applying it to a classical and important problem in machine learning and data science: computing nearest neighbors. The basic problem of computing the nearest
neighbor of a point $x_1$ among $n-1$ points $x_2,\hdots,x_n$ in $d$-dimensional space is costly if both $n$ and $d$ are large, since there are many candidates and each distance computation is expensive.
As an example consider $\ell_1$ distance where $\rho(x_1,x_i) = \sum_{j=1}^d |x_{1,j} - x_{i,j}|$, where $x_{i,j}$ denotes the $j$-th coordinate of point $x_i$. 
Solving this problem exactly requires summing the coordinate-wise distances across all coordinates for each point, which is computationally intensive when $d$ is large.

Casting this problem in the framework of BMO, the set $\mathcal{I}$ to corresponds to the $n-1$ points $x_2, \ldots, x_n$,   and  $\theta_i \triangleq \frac{1}{d} \cdot \rho(x_1,x_i)$, where our goal is to find the point $x_i$ with the smallest $\theta_i$.
Referring to Figure \ref{fig:block_dg}(a), the first step is to construct a Monte Carlo box to efficiently generate unbiased estimates of $\theta_i$.
We see that this can be accomplished by sampling $m$ coordinates $J_1,J_2,\hdots, J_m$ independently and uniformly at random and computing 
\begin{equation}
\vspace{-.1cm}
\hat{\theta}_{i,m} = \frac{1}{m}\sum_{k=1}^m|x_{1,J_k} - x_{i,J_k}|.
% \vspace{-.1cm}
\end{equation}
Note that this estimator is unbiased, as $\E  \{\hat{\theta}_{i,m}\} = \theta_i$.
With the ability to procure noisy but unbiased estimates of $\theta_i$, we can estimate each  $\theta_i$ uniformly to a desired accuracy by choosing $m$ sufficiently large, and output the point with the smallest estimated distance, as in Figure \ref{fig:block_dg}(b). 
However, this wastes computational power on points that are far from $x_1$, as these estimates will with good probability be large after relatively few samples.
This motivates Figure \ref{fig:block_dg}(c), where the Monte Carlo boxes are \textit{adaptively} invoked with an increasing number of samples to find the minimum. Just as in the medoid application, the Monte Carlo estimates can be efficiently updated given additional samples. 

We apply this accelerated nearest neighbor subroutine to obtain a new algorithm, bandit-based Monte Carlo optimization for nearest neighbors (\texttt{BMO-NN}), which provides theoretical guarantees on returning the exact nearest neighbors.
We show in Figure \ref{fig:BMONN_compGainIntro}
that \texttt{BMO-NN} obtains dramatic improvements in practice, allowing
for an 80x reduction in the number of 
coordinate-wise distance computations needed over exact computation on the $100k$ Tiny ImageNet dataset to obtain the exact $k=5$ nearest neighbors with target error probability $\delta= .01$.
Some
popular $k$-NN 
algorithms are based on LSH \cite{andoni2015falconn_LSH}, which provides theoretical guarantees
for returning approximate
\begin{figure}[t]
  \begin{center}
  \includegraphics[width=.6\linewidth, trim = 0 0 0 22, clip]{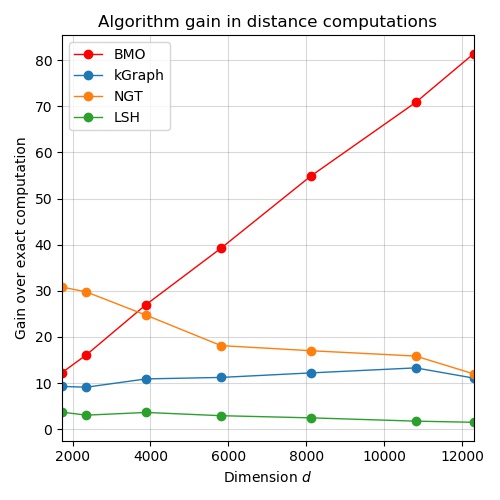}
  \vspace{-.3cm}
  \caption{Gain in number of coordinate-wise distance computations needed by \texttt{BMO-NN} and other algorithms over exact computation on the 100k Tiny ImageNet dataset for $k=5$ exact nearest neighbors at target error probability $\delta= .01$, details in Section \ref{sec:experiments}.}\label{fig:BMONN_compGainIntro}
\end{center}
\vspace{-.3cm}
\end{figure}
% here before
nearest neighbors (while our theoretical guarantees are for exact nearest neighbors). As can be seen in Figure \ref{fig:BMONN_compGainIntro}, LSH requires 50x more distance computations than \texttt{BMO-NN}.
Comparing with heuristic algorithms with no theoretical guarantees, \texttt{BMO-NN} provides a 7x improvement over kGraph \cite{KGraph} and NGT \cite{iwasaki2018optimization}, two of the best performing approximate nearest neighbor algorithms according to evaluations from the benchmark of \cite{ann-benchmarks}.
We prove a bound on the complexity of \texttt{BMO-NN}, and show that under some regularity conditions on the data the complexity of $k$-NN is reduced from $O(nd)$ for exact computation to $O\left((n+d)\log^2 \left(\frac{nd}{\delta}\right)\right)$ for error probability $\delta$ using our algorithm.
This can be seen by the almost linearly increasing gain as a function of $d$ in Figure \ref{fig:BMONN_compGainIntro} for \texttt{BMO-NN}. 

\subsection{Related work on $k$-nearest neighbors}
Nearest neighbors is a fundamental data science primitive, and has seen much recent use in machine learning tasks \cite{shakhnarovich2006nearestNeighborsBook}.
$k$-nearest neighbor graph construction [Chapter 13]\cite{HasTibFri} is a commonly used subroutine in many unsupervised
learning algorithms such as
Isomap \cite{TenSilLan}, locally linear
embeddings \cite{RowLaw}, 
Hessian Eigenmaps \cite{DonGri},
and some spectral clustering algorithms like
\cite{LinMisEtAl}.
In recent years there has been a lot of work on $k$-NN algorithms
for high dimensional image data such as that on 
the YFCC100M dataset in \cite{JohDouHer}. 
In low dimensions there are some heuristic algorithms 
that are known to perform well such as 
$k$-d trees \cite{BentleyKDTrees} and
ball-trees \cite{OmoBallTree}.
Other heuristic algorithms like kGraph \cite{DonChaLi} and NGT \cite{iwasaki2018optimization} can also operate in higher dimensional regimes, creating graph based data structures that allow for fast querying.
These algorithms lack theoretical guarantees, but empirically yield a computational complexity scaling sub-quadratically in $n$ for nearest neighbor graph construction, which naively requires $O(n^2d)$ time, empirically reducing this to approximately $O(n^{1.14}d)$ in \cite{DonChaLi}.

Another common approach is based on locality sensitive hashing (LSH),
with recent algorithms like Falconn \cite{andoni2015falconn_LSH} returning approximate nearest neighbors for certain distance metrics in provably sub-quadratic time in $n$.
LSH schemes generally exhibit a query time complexity of $O(n^{\rho}d)$ to find an approximate nearest neighbor of a single point for an approximation factor dependent  $\rho<1$ (for example, $\rho=1/7$ for $\ell_2$ distance for an approximation factor of 2); \cite{razenshteyn2017high} provides a broader survey of these results.
Note that these methods still require time linear in the dimension $d$ of the data points, which can be extremely costly in practice.
One technique for avoiding high dimensionalities involves preprocessing the data by projecting it into 
lower dimensional space by appealing to the Johnson-Lindenstrauss lemma \cite{JL1984}, %incurring some distortion of pairwise distances,
and then computing $k$-nearest neighbors in this lower dimensional space. 
While dimensionality reduction is often helpful, in applications like
embedding graphs \cite{wang2016structural}
and videos \cite{ramanathan2015learning},  
the dimensions of the space these objects are projected to is still in the thousands, making
$k$-nearest neighbors
a problem of interest in high dimensions.

We focus on high dimensional data in this paper as \texttt{BMO-NN} is designed for computations on raw (un-projected) data. \texttt{BMO-NN} is an alternative to the pipeline of JL followed by a nearest neighbor algorithm on the low dimensional data, achieving comparable complexity while providing theoretical guarantees on returning the exact nearest neighbor (in contrast to JL). Interestingly, \texttt{BMO-NN} can be viewed almost as an adaptive JL;
it is unnecessary to project all points into the same lower dimensional space in a one-shot procedure, by subsampling and measuring coordinate-wise distances we are adaptively projecting pairs of points to the necessary lower dimensional space.
Further, dimensionality reduction techniques are primarily limited to $\ell_2$ distance, where in many biological applications (like the RNA-Seq dataset we consider \cite{10xdata}) one is interested in $\ell_1$ distance, for which there is no good embedding \cite{brinkman2005impossibility}, and so one needs to operate on the raw high dimensional data.

Our bandit-based algorithm operates in a fundamentally different
way from algorithms like kGraph, NGT, and Falconn. 
While kGraph and NGT improve their sample complexity by computing the distances between fewer 
pairs of
points using the fact that the neighborhoods of 
neighboring points have large
intersections
(giving them good scaling with 
the $n$), \texttt{BMO-NN}
improves its sample complexity
by computing distances between all pairs of points
approximately 
(giving it good scaling with  $d$).
For example, \texttt{BMO-NN} does not yield an efficient data structure for finding the nearest neighbor of a new point, and will require looking at all $n$ other points for this task.
Our algorithm also operates differently from projection or hashing based methods like LSH, as it \textit{adapts} to problem difficulty, where LSH computes the same number of hashes for each point.
Further, all three of these algorithms have expensive indices or data structures that need to be pre-computed, which we do not include in our wall-clock or FLOP comparisons, in stark contrast with \texttt{BMO-NN}.

\subsection{More recent applications of BMO technique}

Subsequent to a preliminary version of this manuscript being made public,  
several papers have built on this work and found new applications of the BMO technique. 
\cite{lejeune2019adaptive} built off of this work to efficiently solve a relaxation of the $k$-NN problem, which tries to find a subset of $O(k)$ points that contain the $k$-nearest neighbors of a given query point.
$k$-NN graph construction was later also considered in \cite{mason2019learning,mason2021nearest}, where the authors utilize the triangle inequality to improve scaling with $n$.
\cite{zhang2019adaptive} efficiently solves the problem
of Monte Carlo permutation-based multiple testing using the BMO technique.
\cite{baharav2019Medoid} improves upon \cite{BagKamNtrZhaTse} for the medoid problem by opening up the Monte Carlo boxes to induce and exploit correlation between the arms.
The problem of $k$-medoids clustering is solved using the BMO technique in
\cite{tiwari2020banditPAM}.
A rank-one estimation problem was efficiently solved in \cite{kamath2020Adaptive} using this technique.
\cite{singhal2020query} constructs a BMO-inspired algorithm for mode estimation of a multivariate distribution.

\subsection{Outline} 
 
In Section \ref{sec:mab}, we apply multi-armed bandit theory to develop a specific BMO algorithm based on UCB1, an upper-confidence bound (UCB) algorithm. We 
specialize it to $k$-NN in Section \ref{sec:knn}, describing algorithms for both exact and approximate nearest neighbor computation.
We then discuss possible improvements in cases where the data is
sparse in Section \ref{sec:sparsity}
and when the distance metric is
$\ell_2$ in Section \ref{sec:l2}.
In Section \ref{sec:experiments} we provide simulation results that corroborate our theoretical findings and demonstrate the practicality of our algorithm. 
We discuss possible extensions of the BMO technique and the potential for using more sophisticated bandit algorithms in Section \ref{sec:disc}. Section \ref{sec:conclusions} concludes the paper.
Proofs are largely relegated to the Appendices, which are included as supplementary material.
Appendix \ref{app:knn_sup} provides proofs for the sample complexity and correctness of our algorithm,
Appendix \ref{app:knn_sup_normalproof} proves the $O((n+d)\log^2(\frac{nd}{\delta}))$ sample complexity bound for random Gaussian data,
Appendix \ref{app:MCboxes} provides additional discussion and proofs for our improved Monte Carlo boxes in Section \ref{sec:improvedMCboxes}, and Appendix \ref{app:details_of_experiments} provides simulation details.

\section{Multi-armed bandits for BMO}
\label{sec:mab}

\subsection{Background}\label{sec:related}

In the field of multi-armed bandits, the pure exploration setting has received a surge of recent interest \cite{ audibert2010best, jamieson2014best, kaufmann2016complexity}. Of primary interest to us is the top-$k$ problem, where one tries to find the $k$ arms with the largest means, trading off between the number of samples needed and the error probability
\cite{chen2017nearly, kalyanakrishnan2012pac, simchowitz2017simulator}.
There are many sophisticated algorithms
developed to solve this problem and the best-arm identification problem, some of which are provably optimal \cite{jamieson2014lil}.  
We
demonstrate the power of using bandits as a tool by building off of the simple UCB1 algorithm of \cite{auer2002finite}, showing that the power of the BMO technique comes from the reduction of a computational problem to one of adaptive statistical estimation, and not from a specialized bandit algorithm.
We note that modifying 
more sophisticated bandit algorithms
can further improve our theoretical bounds, and discuss this in Section \ref{sec:beyondUCB}.

A similar problem has also been studied under the framework of simulation optimization (SO).
In this setting, motivated by the optimization of queueing systems and supply chain inventory, the objective is to maximize a function $J(x)$ over a design parameter $x$, where $J(x)=\E \{f(x;\xi)\}$ is expressible as an expectation.
Early works on SO include \cite{ho1992ordinal}, which showed that if $x$ can only take values in a finite set, ranking these designs is potentially an easier problem than estimating all of their values.
For maximizing $J(x)$ over a large discrete set of $x$'s, \cite{chen2000simulation} built algorithms off of the same principle as that in the multi-armed bandit literature; that better candidates should be sampled more to reduce the probability of error in identifying the best $x$. 
SO has been the subject of many more recent studies, including works on heavy-tailed distributions \cite{glynn2004large}, finite sampling budgets \cite{shin2018tractable}, and other variants \cite{kim2006selecting}, including works on ranking and selection \cite{kim2007recent}.
Most works however deal with a more complicated setting than ours, where the distributions are unknown, and so the primary focus is to obtain the optimal asymptotic behavior \cite{kim2006asymptotic}.
Additionally, due to the underlying computational problem in our scenario, we can exactly compute the value of a given design, a feature not present in the standard SO literature, as in our problem stochasticity is artificial and generated by our procedure, where in SO it is inherent to the model.
For a more recent work detailing further SO literature and its relation to stochastic multi-armed bandits, we refer the reader to \cite{glynn2015ordinal}.

\subsection{The modified UCB1 algorithm \texttt{BMO UCB}} \label{sec:modified_UCB_alg}
We now describe a UCB based bandit sampling algorithm for BMO.
This routine takes in a set of arms $\{a_{i}: i \in[n]\}$ (Monte Carlo boxes) with associated methods for pulling them, including how to update
the estimates of their mean and construct confidence
intervals around these estimates.
Further, to account for the underlying computational problem, this routine also takes as input a method for evaluating the mean of
an arm exactly, which is computationally expensive. 
With the true value (mean) of arm $i$ being $\theta_i$, let $\hat{\theta}_{i, \ell}$
be the estimate of the mean of arm $i$ after $\ell$ pulls (samples), and $2C_{i,\ell}$ be the width of the confidence interval for $\hat{\theta}_{i,\ell}$ after $\ell$ pulls.
Further, let $T_i(t)$ denote the number of times arm $i$ has been pulled prior to time $t$.
\begin{algorithm}[t]
  \caption{\texttt{BMO UCB} \label{alg:genUCB}}
\begin{algorithmic}[1] 
    \State \textbf{Input: } $\{a_{i}: i \in [n]\}, \sigma, \texttt{MAX\_PULLS}, k,\delta$
    \State $t\gets 1$ \Comment{iteration counter}
    \State $B \gets \emptyset$ \Comment{set of $k$ best arms}
    \State $S \gets [n]$ \Comment{set of arms under consideration}
    \State Pull each arm once
    \While {$|B| < k$} 
    \State Compute
  $I_t = \underset{i\in S}{\argmin}\ \hat{\theta}_{i, T_i(t)}-C_{i,T_{i}(t)}$
  \If {$\hat{\theta}_{I_t, T_{I_t}(t)} + C_{I_t,T_{I_t}(t)} < \hspace{-.3cm}\underset{i\in S, i \neq I_t}{\min} \hat{\theta}_{i, T_i(t)} - C_{i,T_i(t)}$ \label{alg:line:addToSetCond}}
  \State Add $I_t$ to $B$ and remove $I_t$ from $S$ \label{alg:line:addBest}
  \State Continue to next iteration of loop 
  \EndIf
  \If {$T_{I_t}(t) < \texttt{MAX\_PULLS}$} 
  \State Pull arm $a_{I_t}$ 
  \State Construct $\hat{\theta}_{I_t, T_{I_t}(t+1)}$ and $C_{I_t,T_{I_t}(t+1)}$
  \Else 
  \State Evaluate the mean of arm $a_{I_t}$ exactly \label{alg:line:bruteForce}
  \State Set $\hat{\theta}_{I_t, T_{I_t}(t+1)} = \theta_{I_t}$, 
  $C_{I_t,T_{I_t}(t+1)}=0$. 
  \EndIf
  \State $t \gets t + 1$
  \EndWhile
\end{algorithmic}
\end{algorithm}

\texttt{BMO UCB} (Algorithm \ref{alg:genUCB}) is essentially UCB1 \cite{auer2002finite},
with the added condition that if an arm
is pulled more than \texttt{MAX\_PULLS} times
we evaluate its mean exactly (line 13), which is only possible due to the underlying computational problem. 
Prior to this threshold of \texttt{MAX\_PULLS} samples, we simply query our Monte Carlo boxes for unbiased samples; in the case of $k$-NN, they sample coordinate-wise distances with replacement.
If an arm is selected to be pulled that has already been pulled \texttt{MAX\_PULLS} times, our algorithm decides that enough samples have been expended trying to approximate its mean, and we should instead exactly evaluate the mean of this arm from scratch; for $k$-NN this means summing all $d$ coordinate-wise distances, at the cost of $d$ additional samples.
This hybrid procedure of sampling (\textit{with replacement}) until an arm is deemed close enough to one of the best arms that its mean needs to be exactly evaluated (in the case of $k$-NN via sampling \textit{without replacement}) is in fact order optimal, as discussed in the next subsection.

% This yields a total cost of $2\cdot$\texttt{MAX\_PULLS}, at most a factor of 2 worse than if our Monte Carlo boxes sampled without replacement, while allowing for the computational efficiency of sampling with replacement.

% \blue{
While more sophisticated bandit algorithms like lil'UCB \cite{jamieson2014lil} can be used for best-arm identification, out of the box UCB1 cannot, as the number of times the best arm needs to be pulled is unbounded.
With \texttt{BMO UCB} however, we know that the best arm cannot be pulled more than \texttt{MAX\_PULLS} times before its confidence interval shrinks to 0 and the algorithm terminates.
This algorithm can be implemented efficiently by maintaining a priority queue on $\hat{\theta}_{i,T_i(t)}-C_{i,T_i(t)}$, requiring only $O(\log n)$ computational overhead per iteration (as only a single arm is queried and updated).
This algorithm takes as input the set of arms, a bound on the sub-Gaussian parameter of the estimators $\sigma$, the integer  \texttt{MAX\_PULLS}, and the
number of best arms to be returned $k$.
%}

\subsection{Confidence intervals}\label{sec:knnDetails}

In order to analyze \texttt{BMO UCB}, we want to make statements of the form $\P\left(\left|\hat{\theta}_{i, T_i(t)}-\theta_i\right| \ge C_{i,T_i(t)}\right) \le \delta'$.
This means we want to know how fast our estimators $\hat{\theta}_{i,m}$ concentrate.
For simplicity, we assume for the rest of this paper that our Monte Carlo estimates are averages of independent samples.
One natural case where this occurs is when the Monte Carlo estimates of $\theta_i$ are the sum of independent samples of a carefully constructed random variable $X_i$, as is the case in $k$-NN which is the main focus of this paper.
Here $\theta_i = \frac{1}{d}\sum_{j=1}^d |x_{1,j}-x_{i,j}|$ and $X_i=|x_{1,J} - x_{i,J}|$ for $J\sim\text{Unif}([d])$.
The Monte Carlo estimator we used was $\hat{\theta}_{i,m} = \frac{1}{m}\sum_{k=1}^m|x_{1,J_k} - x_{i,J_k}|$ for $J_k\underset{i.i.d.}{\sim}\text{Unif}([d])$, the mean of $m$ independent samples from $X_i$. 
For the rest of this work we additionally assume that these $X_i$ are sub-Gaussian random variables, and discuss loosening these assumptions in Section \ref{sec:disc}.

Given that our estimates are sums of independent samples of a sub-Gaussian distribution, we are able to utilize Hoeffding's inequality to construct confidence intervals on $\hat{\theta}_{i,m}$.
Recall that a random variable $X$ with mean $\mu = \E \{X\}$ is said to be sub-Gaussian with parameter $\sigma$ if $\E \left\{e^{\lambda (X-\mu)}\right\} \le e^{\sigma^2\lambda^2/2}$ $\forall \lambda \in \R$.
We utilize the Orlicz norm of a random variable $\| \cdot\|_{\Psi_2}$ which denotes the minimum valid sub-Gaussian constant of a random variable.
Given $\sigma_i \ge \| X_i\|_{\Psi_2}$, we can then construct confidence intervals as
\begin{equation} \label{eq:CI}
\hspace{-.15cm}C_{i,T_i(t)} = \begin{cases}
 \sqrt{\frac{2 \sigma_i^2 \log \frac{2}{\delta'}}{T_i(t)}} & \text{ if } T_i(t) \le \texttt{MAX\_PULLS}\\
 0 &\text{ if } T_i(t) > \texttt{MAX\_PULLS}
 \end{cases}
\end{equation}
noting that our confidence intervals will depend on both a bound $\sigma_i$ on the sub-Gaussian constant of $X_{i}$ and a confidence $\delta'\in(0,1)$.
In the following lemma we show that these are valid confidence intervals.

\begin{lemma}\label{lem:conf_intervals}
With probability at least $1- \delta$ each $\theta_i$ lies within its $\sigma_i \ge\|X_i\|_{\Psi_2} ,\ \delta'=\frac{\delta}{n\cdot \texttt{MAX\_PULLS}}$ confidence intervals during the entirety of \texttt{BMO UCB}.
\end{lemma}
\begin{proof}
We observe that if arm $i$ has been pulled fewer than $\texttt{MAX\_PULLS}$ times at
time $t$, then $T_i(t)$ is equal to the number 
of times the arm is selected by \texttt{BMO UCB}.
Then $C_{i,T_i(t)}$ is a valid $(1-\delta')$-confidence interval by Hoeffding's inequality.
If arm $i$ has been pulled $\texttt{MAX\_PULLS}$ times previously and is selected again, then the arm mean is exactly computed (line \ref{alg:line:bruteForce} of Algorithm \ref{alg:genUCB}).
Since our estimated arm mean will now be the true arm mean, $C_{i,T_i(t)} = 0$ holds with probability 1.
Noting that each arm will be pulled at most $\texttt{MAX\_PULLS}$ times, at most
$n\cdot\texttt{MAX\_PULLS}$ arms pulls are made by $\texttt{BMO UCB}$.
Since only one confidence interval is constructed in each iteration, failing with probability $\delta' = \frac{\delta}{n\cdot\texttt{MAX\_PULLS}}$, we have by a union bound that with probability at least $1-\delta$ the true arm means
are always within their confidence intervals.
\end{proof}

Note that for a given problem instance the $X_i$'s are bounded random variables. For example, in the case of $k$-NN for $\ell_2$ distance, $|X_i| \le \max_{i,j \in [n],\ell \in [d]} (x_{i,\ell}-x_{j,\ell})^2$ with probability 1 for all $i$.
This implies that the estimators used in \eqref{eq:knn1} are indeed sub-Gaussian random variables.
While assuming knowledge of each of these sub-Gaussian constants $\|X_i\|_{\Psi_2}$ is impractical, we require an upper bound on them to construct confidence intervals to run \texttt{BMO UCB}, noting that other algorithms can make alternative assumptions \cite{audibert2006use,cowan2017normal}.
In practice we do not know these sub-Gaussian constants, and instead estimate a global $\sigma$ for all arms from a few initial samples and update it after every pull.
We discuss improving these constants in Section \ref{sec:improvedMCboxes}.

One of the primary goals of this paper is to demonstrate the general applicability of the BMO method. 
In many applications, the expensive to compute but easy to approximate quantities (the $\theta_i$ of \eqref{eq:prob_formulation_intro}) are sums, and so a natural questions is whether constructing Monte Carlo boxes that sample without replacement significantly outperform our hybrid ones that primarily sample with replacement.
If our Monte Carlo boxes were constructed by sampling \textit{without} replacement, then our confidence intervals for our estimators would be multiplied by the square root of the finite population correction factor $\texttt{fpc}\triangleq (1-T_i(t)/\texttt{MAX\_PULLS}) $. 
Such a change would not impact the algorithm's orderwise sample complexity however, as for $T_i(t)<\texttt{MAX\_PULLS}/2$ we have \texttt{fpc}$>1/2$, meaning that a confidence interval of the same width can be constructed by sampling with replacement using at most double the number of samples.
For $T_i(t)\ge \texttt{MAX\_PULLS}/2$, we see that since our scheme samples an arm no more than $2\cdot \texttt{MAX\_PULLS}$ times, we use at most four times the number of samples in this regime.
Hence, our hybrid scheme requires at most 4x as many samples as sampling without replacement.
Since sampling without replacement requires additional storage and complexity, storing the sampled indices and computing a random point that hasn't yet been sampled in every iteration, we focus on Monte Carlo boxes that sample with replacement for simplicity's sake.

\section{The \texttt{BMO-NN} algorithm}\label{sec:knn}
In this section we reformulate the $k$-nearest 
neighbor problem using the BMO technique, 
and provide theoretical guarantees on the performance of the algorithm 
we obtain.
Consider $n+1$ points, $x_0, \hdots, x_n$ $\in \ \mathbb{R}^d$. For the sake of concreteness 
we focus the rest of the exposition in this paper on finding
the $k$ nearest neighbors of $x_0$ under $\ell_2$ distance,
but the approach works identically for general separable distance functions $\rho(x,y) = \sum_{j=1}^d \rho_j(x_j,y_j)$ where $\rho_j : \R \times \R \mapsto \R$ for $j\in[d]$, e.g. $\ell_p$ distances.
Note that $\rho$ does not need to be a distance metric.
We consider
each of the points $\{x_1, \hdots, x_n\}$ as arms.
Note that the $k$-nearest neighbors
under $\ell_2$ distance are the same 
as those under squared $\ell_2$ distance, which is not a metric. We consider $\theta_i = \frac{1}{d}  \sum_{j=1}^{d} (x_{0,j}- x_{i,j})^2$,
with the objective being to find the $k$ 
points with the 
smallest $\ell_2^2$ distances to $x_0$.
We see that for $J$ sampled uniformly at random from $[d]$ we have that 
$X_i = (x_{0,J}- x_{i,J})^2$ gives an unbiased estimate of $\theta_i$. We can construct an $\ell$ sample estimate of $\theta_i$ by sampling $J_1,\hdots, J_{\ell}$ independently and uniformly at random from $[d]$ and computing
\vspace{-.15cm}\begin{equation}\label{eq:knn1}
     \hat{\theta}_{i, \ell} = \frac{1}{\ell} \sum_{m=1}^{\ell} 
     (x_{0,J_m}- x_{i,J_m})^2.\vspace{-.05cm}
\end{equation}
To update our $\ell-1$ sample estimator after the $\ell$-th pull, we can perform 
% \begin{align}
    $\hat{\theta}_{i, \ell} = \frac{1}{\ell} \left(  (\ell-1) \hat{\theta}_{i, \ell-1}
    + (x_{0,J_{\ell}}-x_{i, J_{\ell}})^2 \right),$
% \end{align}
which takes $O(1)$ time. 
Further, $\theta_i$ can be exactly computed in $d$ samples, giving us \texttt{MAX\_PULLS}$=d$.
This reformulates $k$-NN into the BMO framework, with the arms as the random variables $X_i = (x_{0,J}-x_{i,J})^2$ for $J \sim \text{Unif}\left([d]\right)$, where solving
this multi-armed bandit problem 
gives us the $k$-nearest neighbors of $x_0$.

By carefully constructing our Monte Carlo boxes to align with the $k$-NN objective, we are able to leverage \texttt{BMO UCB} and create \texttt{BMO-NN} (Algorithm \ref{alg:UCB-knn}).
\texttt{BMO-NN} iterates over each arm $j\in [n]$ and finds its nearest neighbors using \texttt{BMO UCB}.
\texttt{BMO UCB} is a general algorithm which can be utilized to solve other problems by tailoring problem specific Monte Carlo boxes (methods to construct confidence intervals and unbiased estimators).

\begin{algorithm}[h]
  \caption{\texttt{BMO-NN} \label{alg:UCB-knn}}
\begin{algorithmic}[1]
  \State \textbf{Input:} $x_1,\hdots x_n \in \R^d,\ \sigma,\ k,\ \delta$ 
  \For{$i =1,\hdots,n$ }
    \State \parbox[t]{195pt}{\hangindent=.5cm Construct arms $\left\{a_{j}\right\}_{j=1, j\neq i}^{n}$
    with estimators as in \eqref{eq:knn1}
    and confidence intervals as in \eqref{eq:CI} \strut}
    \State $k$-NN of $x_i \; \gets $   {\texttt{BMO UCB}}\big($\left\{a_{j}\right\}_{j\neq i},\sigma, d,k,\frac{\delta}{n} \big)$
  \EndFor
\end{algorithmic}
\end{algorithm}
\vspace{-.5cm}

\subsection{Exact Nearest Neighbors}

We now analyze the performance of the \texttt{BMO-NN} algorithm.
We let $(\cdot)$ be a permutation on $[n]$ such that under $(\cdot)$ the points are sorted by increasing distance to $x_0$, that is $\theta_{(1)}\le \hdots \le \theta_{(k)} < \theta_{(k+1)} \le \hdots \le \theta_{(n)}$.
We define the set of $k$-nearest neighbors of $x_0$ as $\{x_{(i)} : i\in[k]\}$ and so must have $\theta_{(k)} < \theta_{(k+1)}$ for the set of $k$-nearest neighbors to be well defined.
This is without loss of generality, as if $\theta_{(k)}=\theta_{(k+1)}$, our algorithm can be used to find $k$ of the smallest $\theta_i$.
With this, we define for point $x_0$ the gaps $\Delta_{(i)} =  \theta_{(i)}-\theta_{(k)}$. 
Note that this is the gap between $\theta_i$'s not $\rho(x_0,x_i)$'s, where for $\ell_2^2$ we have $\theta_i = \frac{1}{d} \rho(x_0,x_i)^2$.
We use $x \wedge y$ to denote $\min(x,y)$ and $x \vee y$ to denote $\max(x,y)$.
With this notation formalized, we are now able to state the following theorem
regarding the data dependent performance of \texttt{BMO-NN}.
For clarity \texttt{BMO-NN} and \texttt{BMO UCB} only take as input a universal bound $\sigma$ on the sub-Gaussian parameters of the individual arms, the $\sigma_i$'s, but the results we prove allow the algorithms to take as input the $\sigma_i$'s and utilize these individual bounds.
Note that even if the algorithm fails it will not take more than $2nd$ coordinate-wise distance computations to terminate.

\begin{thm}[Main Theorem]\label{thm:knn1}
Assume \texttt{BMO-NN} receives as input $\{x_i\}_{i=0}^n$, $\{\sigma_i\}_{i=1}^n$ where $\sigma_i \ge \|X_i\|_{\Psi_2}$, $k \in [n]$, and $\delta\in (0,1)$.
Then with probability at least $1-\delta$ \texttt{BMO-NN} returns the correct
$k$-nearest neighbors of $x_0$, requiring on this success event
\begin{equation*} \label{eq:knn_proof_statement}
M
\le 2kd+\sum_{i = k+1}^{n} \left( \left\lceil \frac{8 \sigma_{(i)}^2}{\Delta_{(i)}^2} \log \left( \frac{2nd}{\delta} \right) \right\rceil  \wedge 2d \right)
\end{equation*}
coordinate-wise distance computations.
\end{thm}
% \vspace{-.5cm}
\begin{proof}
The details of the proof of this Theorem are relegated to Appendix \ref{app:knn_sup}.
At a high level, we proceed by assuming that all arm mean estimates stay within their confidence intervals, and bound the number of times each suboptimal arm is pulled.
This is done by noting that once an arm's confidence interval is sufficiently small (half its gap to the best arm) it will never be pulled again.
Since we know the rate at which confidence intervals decay, we can bound the number of times each suboptimal arm is pulled, noting that if the gap is too small the algorithm can exactly evaluate the arm mean, requiring $2d$ total distance computations.
\end{proof}
\vspace{-.2cm}
We can compare our bound in \eqref{eq:knn_proof_statement} to the sample complexity of identifying the top-$k$ arms in the standard multi-armed bandit setting, where the sample complexity is shown in \cite{simchowitz2017simulator} to be lower bounded by
\begin{equation} \label{eq:topKLowerBound}
    \sum_{i=1}^k \frac{\sigma_i^2\log\left(\frac{1}{\delta}\right)}{(\theta_{(k+1)}-\theta_{(i)})^2} + \sum_{i=k+1}^n \frac{\sigma_i^2\log\left(\frac{1}{\delta}\right)}{(\theta_{(i)}-\theta_{(k)})^2}
\end{equation}
assuming arm pulls are independent across arms and time, and that pulls from arm $i$ are distributed as $\CN(\theta_i,\sigma_i^2)$.
One immediate difference that can be seen between the expression in Theorem \ref{thm:knn1} and \eqref{eq:topKLowerBound} is that the sample complexity in Theorem \ref{thm:knn1} has a minimum with $2d$ taken for each term.
This is due to the underlying computational problem in our setting; in the standard multi-armed bandit setting, if two arms have means $\epsilon$ apart, we need to take on the order of $\epsilon^{-2}$ samples to determine which is smaller with error probability nontrivially below $1/2$.
However in our scenario of $k$-NN, we recall that there is an underlying computational problem; these $\theta_i$ are actually distances between two points, which we can easily approximate with our Monte Carlo boxes, but can also exactly compute in $d$ time.
This means that if two arms have very similar means, we can simply evaluate the two arm means exactly by computing the distances between the two pairs of points, requiring computation independent of $\epsilon$. 
This makes standard best-arm identification algorithms uncompetitive when sampling with replacement.
We additionally see that the second sum in \eqref{eq:topKLowerBound} matches our theorem aside from the minimum when ignoring log factors, but the first $k$ terms do not.
This is due to the suboptimality of UCB1 as an exploration algorithm, but can be improved as discussed in Section \ref{sec:beyondUCB}.

While Theorem \ref{thm:knn1} is phrased for finding the nearest neighbor of $x_0$, we can replace $x_0$ with any other point $x_j$ in our dataset and the theorem will still hold, albeit with the $\Delta_i$'s and $\sigma_i$'s dependent on $x_j$ rather than $x_0$.
Theorem \ref{thm:knn1} is general, holding for arbitrary gaps and $\sigma_i$'s.
We provide the following Proposition, evaluating the sample complexity in Theorem \ref{thm:knn1} in the specific case where $\theta_i \underset{i.i.d.}{\sim} \CN(\mu,1)$, to provide intuition for how our sample complexity compares to that of exact computation.
While the mean of the normal distribution does not show up explicitly in the result, to ensure that all distances are positive with high probability we require that $\mu= \Omega\left(\sqrt{2\log n } \right)$.
The proof of this Proposition is deferred to Appendix \ref{app:knn_sup_normalproof}.
\begin{prop}[Complexity under Gaussian means] \label{cor:knn_normal}
Assume that a dataset $x_0,\hdots, x_n \in \R^d$ is randomly generated such that $\theta_i \underset{i.i.d.}{\sim} \mathcal{N}(\mu,1)$ with $\mu= \Omega\left(\sqrt{2\log n } \right)$ and $\|X_i\|_{\Psi_2}\le \sigma$ for all $i \in [n]$ with constant $\sigma$.
Then, if \texttt{BMO-NN} receives as input $\{x_i\}_{i=0}^n$, $\sigma$, integer $k>0$ with $k\le \frac{1}{6}\log^{1/2-c'} (n)$ for any constant $c'\in(0,1/2)$, and $\delta\in (0,1)$,
then with probability at least $1-\delta$ \texttt{BMO-NN} will return the correct
$k$-NN of $x_0$ using in expectation (with respect to the random $\theta_i$)
\begin{equation*}
    O\left( \left(n+d\right)\frac{\log^2 \left( \frac{nd}{\delta} \right)}{\log^2 \log \left( \frac{nd}{\delta} \right)} \right)
\end{equation*}
coordinate-wise distance computations.
\end{prop}
Extending the assumptions of Proposition \ref{cor:knn_normal} to other points and distances gives us that \texttt{BMO-NN} 
succeeds in $k$-NN graph construction with probability at least $1-\delta$ and requires $O\left(n\left(n+d\right)\log^2 \left(\frac{n^2d}{\delta}\right)\right)$ coordinate-wise distance computations in expectation 
over randomness in the algorithm and the distances $\{\rho(x_i,x_j) : i,j\in[n]\}$.
This yields a better dependence on the dimension $d$ than the linear $O(n^2d)$ of exact computation, but retains the quadratic dependence on $n$ due to the fact that we approximate the distance between all pairs of points.

As noted before, each coordinate-wise distance computation requires only $O(\log n)$ computational overhead in maintaining the priority queue of arms, and so we can translate sample complexity bounds for \texttt{BMO-NN} to running time ones with only an additional $\log n$ factor.
\subsection{Approximate Nearest Neighbors}
While finding the $k$ nearest neighbors of a point is a useful primitive, there are some scenarios where this may be unnecessarily expensive. For example, if there are many points only slightly further than the $k$-th nearest neighbor, a lot of work will need to be done to separate them, whereas in practice any of these points would have been ``close enough''.
In these scenarios, one may simply want to return with probability at least $1-\delta$ a set of $k$ points all of which have distance at most $\epsilon$ greater than that of the true $k$-th nearest neighbor.
This setting is commonly referred to as the PAC (Probably Approximately Correct) formulation \cite{even2002pac}, or the indifference-zone formulation in the Simulation Optimization community \cite{kim2001fully}.
Converting \texttt{BMO-NN} to an additive PAC algorithm which we refer to as PAC \texttt{BMO-NN} requires only one minor modification; changing line \ref{alg:line:addToSetCond} of \texttt{BMO UCB} to add the selected arm to the output set when either its confidence interval separates from the others as is currently written, or when its confidence interval has width less than $\epsilon/2$.
With $\Delta_i$ as before we can state the following theorem, proved in Appendix \ref{sec:knn_sup_PAC},
showing that our sample complexity bound, like in Theorem \ref{thm:knn1}, takes a similar form to that of specialized stochastic multi-armed bandit algorithms for the PAC setting \cite{kalyanakrishnan2012pac}.

\begin{thm} [Additive PAC formulation]\label{thm:knn_approx_sec}
Assume PAC \texttt{BMO-NN} receives as input $\{x_i\}_{i=0}^n$, $\{\sigma_i\}_{i=1}^n$ where $\sigma_i \ge \|X_i\|_{\Psi_2}$, $k \in [n]$, and $\delta\in(0,1)$.
Then with probability at least $1-\delta$ PAC \texttt{BMO-NN} returns the correct
$k$-nearest neighbors of point $x_0$ up to 
an additive $\epsilon$ approximation for each, requiring on this success event
\begin{equation*} \label{eq:knn_approx_pulls}
M \le 2kd+\sum_{i=k+1}^{n}   \left( \left\lceil\frac{8 \sigma_i^2}{\left(\Delta_{(i)} \vee \epsilon\right)^2} \log \left(\frac{2nd}{\delta}\right)\right\rceil \wedge 2d\right) 
\end{equation*}
coordinate-wise distance computations.
\end{thm}

To better understand the number of coordinate-wise distance computations required in Theorem \ref{thm:knn_approx_sec}, we examine the case of $k=1$ and consider a power law distribution on the gaps (this was previously done for the best-arm identification setting in \cite{jamieson2013finding}).
Whereas in the case of exact $k$-NN we assumed a normal distribution on the arm means to give a simple expected runtime, here we assume a power law distribution on the gaps. This is because if the means are normally distributed, there will be only a constant number of points within any constant $\epsilon$ of the minimum (with high probability) and so finding an epsilon best arm effectively degenerates to finding the best arm.

\begin{cor} [PAC complexity under power law distributed gaps]
Assume that a dataset $x_0,\hdots x_n \in \R^d$ is randomly generated such that $\Delta_i \underset{i.i.d.}{\sim} \Delta$, where $F(\Delta) = \Delta^\alpha$ for $\Delta \in [0,1]$, with constant $\alpha\in[0, \infty)$.
Given this dataset, constant $\sigma\ge \|X_i\|_{\Psi_2}$ for all $i\in [n]$, and $\epsilon>d^{-1/2}$ with $d=O(n)$, PAC \texttt{BMO-NN} with $k=1$ will, with probability at least $1 - \delta$, identify a point $x_i\neq x_0$ such that $\theta_i \le \min_{j \in [n]} \theta_j + \epsilon$, requiring $M$ coordinate-wise distance computations on this success event where, taking the expectation with respect to these $\Delta_i$, we have
\begin{equation}
    \E\{M\} \le 
    \begin{cases} 
      O\left(n\log \left(\frac{nd}{\delta}\right) \epsilon^{\alpha-2}\right) & \alpha \in [0,2) \\
      O \left(n \log \left(\frac{nd}{\delta}\right) \log \frac{1}{\epsilon} \right) & \alpha = 2 \\
      O \left(n \log \left( \frac{nd}{\delta} \right) \right) & \alpha > 2 
   \end{cases}
\end{equation}
\end{cor}
\begin{proof}
Using Theorem \ref{thm:knn_approx_sec} and
integrating out $\Delta_i$ we see that $\E \{M\}=$
\begin{align*}
O \bigg( n\log \left(\frac{nd}{\delta}\right) \epsilon^{\alpha-2} + n\log \left(\frac{nd}{\delta}\right) \int_{\epsilon}^1 \hspace{-.25cm}\Delta^{\alpha-3}d\Delta\bigg).
\end{align*}
Evaluating this for different $\alpha$ yields the result.
\end{proof}

We observe a clear gradual transition as we vary $\alpha$.
We compare this with the sample complexity of a non-adaptive method, which requires roughly $O\left(n\epsilon^{-2}\right)$ coordinate-wise distance measurements to find an $\epsilon$ best neighbor, ignoring $\delta$ dependence.
Ignoring $\log$ factors, we see that when $\alpha=0$ the two algorithms (adaptive and non adaptive) have the same performance, as all the gaps will be small.
When $\alpha$ is small (less than 2), the gain of an adaptive algorithm is small, $n\epsilon^{\alpha-2}$ as opposed to $n\epsilon^{-2}$ of nonadaptive, since there are many arms within $\epsilon$ of the best.
As $\alpha$ increases the gains get progressively larger;
for $\alpha>2$, we see that the number of pulls required becomes \textit{independent} of $\epsilon$.
This is because there are in expectation only $n\epsilon^{\alpha}$ arms within $\epsilon$ of the best one, so committing $\epsilon^{-2}$ pulls to these difficult ones is relatively cheap, leading to the surprising overall sample complexity.
Note that $\alpha$ is measuring the polynomial rate of decay of the tail of $\Delta$, and so our normally distributed example with exponentially decaying tails falls fully in the $\alpha>2$ case, where the sample complexity is independent of $\alpha$.
\section{Improved Monte Carlo Boxes} \label{sec:improvedMCboxes}
So far we have discussed how to utilize the natural Monte Carlo box for the $k$-NN problem: each $\theta_i$ is the average of $d$ elements, and to get an unbiased estimate of $\theta_i$ we randomly sample a term in this sum. 
Under additional assumptions however, one can design improved estimators of $\theta_i$ that concentrate faster. Here we discuss two approaches, one regarding sampling according to a particular distribution on the $d$ elements, and one on transforming the $d$ elements themselves.
\subsection{Sparse datasets} \label{sec:sparsity}
\texttt{BMO-NN} achieves gains by adaptively sampling
a fraction of coordinates to obtain estimates
of pairwise distances.
Often however, real world datasets are sparse,
in which case the relevant exact computation baseline computes the distance between 2 points in time 
proportional to their sparsity rather than the ambient dimension $d$. 
Ideally we would 
want to design Monte Carlo boxes that 
leverage the sparsity in the dataset.
For example, in the single cell
gene dataset of \cite{10xdata}, only 
$7\%$  of the entries are non-zero, 
even though the points live in $28k$
dimensions. 

For the rest of the section, we focus on the problem of finding nearest neighbors under $\ell_1$ distance, but note that this approach works for any separable distance function where the coordinates that are zero in both points
do not contribute to the 
distance. 
We choose $\ell_1$ for exposition as opposed to $\ell_2$ due to the natural relationship between $\ell_1$ distance and sparse data.
The estimator (Monte Carlo box) we previously proposed for $\theta_i = \frac{1}{d} \|x_0-x_i\|_1$  was 
$X_{i} = |x_{0,J}-x_{i,J}|,$
where $J$ is sampled uniformly at 
random from $[d]$.
However, this estimator is inefficient in that the majority of samples we obtain from it will be 0 if the data is sparse.
Rather than sampling over all coordinates,
we would like to sample coordinates which 
are non-zero in either one of the two points.
We see that by letting $S_0$ and $S_i$ denote the
support of $x_0$ and $x_i$ respectively, with $n_0=|S_0|$ and $n_i=|S_i|$, we have that
\begin{equation*}
    \|x_0-x_i\|_1 = \sum_{j \in [d]} |x_{0,j} - x_{i,j}|
    = \sum_{j \in S_0 \cup S_i} |x_{0,j} - x_{i,j}|.
\end{equation*}
However, computing this set union
of $S_0$ and $S_i$ takes $O(N)$ time where $N=n_0+n_i$,
the time required to compute the exact $\ell_1$ distance between the two, 
neutralizing any gain obtained from our framework. 
We circumvent this expensive operation by 
using the following estimator
which does not explicitly compute the set union:
\begin{equation*}
X_i^\texttt{S}\hspace{-.05cm}= \hspace{-.08cm}
\begin{cases}
\hspace{-.05cm}\frac{N }{2d} \left|x_{0,t^{(0)}}\hspace{-.01cm}-\hspace{-.01cm}x_{i,t^{(0)}}\hspace{-.03cm}\right| (1+ \mathds{1}\{t^{(0)} \hspace{-.05cm} \not\in S_i\}) & \text{w.p. } \frac{n_0}{N}\\
\hspace{-.05cm}\frac{N}{2d}\left|x_{0,t^{(i)}}-x_{i,t^{(i)}}\right|
    (1+ \mathds{1}\{t^{(i)} \hspace{-.05cm} \not\in S_0\}) & \text{w.p. } \frac{n_i}{N}\\
\end{cases}    
\end{equation*}
where
$t^{(0)}$ and $t^{(i)}$ are sampled uniformly at random from $S_0$ and $S_i$ respectively.
Each sample $X_i^\texttt{S}$
is unbiased and 
can be computed in $O(1)$ time,
assuming that one can sample an index uniformly at random from $S_0,S_1$ in constant time.
We formalize this constraint and prove that our Monte Carlo box $X_i^\texttt{S}$ is unbiased in 
Appendix \ref{app:sparsity}.
With this new Monte Carlo box in hand, we can now construct more efficient
estimators for sparse data.
In this scenario, we are able to give the following lemma regarding the improvement afforded by $X_i^\texttt{S}$.

\begin{lemma}\label{lem:sparseEst}
The sparse Monte Carlo box $X_i^\texttt{S}$ satisfies $\sg{X_i^\texttt{S}}\le \frac{n_0 + n_i}{d}\max_{j\in [d]}|x_{0,j}-x_{i,j}|$, where the dense Monte Carlo box has $\sg{X_i} \le \frac{1}{2}\max_{j\in [d]} |x_{0,j}-x_{i,j}|$.
\end{lemma}
This constitutes a reduction in our sub-Gaussian bound by a factor of $\frac{d}{2(n_0+n_i)}$, scaling linearly with the sparsity as desired.
Since the sample complexity of \texttt{BMO-NN} scales as $\sigma_i^2$, we see that if each point has a sparsity of 7\% as in the single cell dataset of \cite{10xdata}, our sample complexity can be reduced by a factor of 12.

While as presented this idea is specific to sparse datasets, it suggests an interesting concept; if our $\theta_i$ are sums, we do not need to sample them uniformly.
Instead, we can re-weight our sampling of the terms in the sum to yield a better estimator, sampling larger terms with higher probability. 
This is analogous to the concept of leverage score based sampling, for example in the application of approximate matrix multiplication, where rows are sampled proportional to their importance in the spectral decomposition of the matrix \cite{mahoney2011_leverage}.
Concretely, we see that if we have $\theta_i = \sum_{j=1}^d \frac{1}{d} z_{i,j}$, we can view this as $\theta_i= \sum_{j=1}^d p_j\cdot\frac{z_{i,j}}{dp_j}$ for any probability distribution $\boldsymbol{p}$ on the $d$ elements.
To generate a Monte Carlo box for this, we let our sample be $X_i=\frac{z_{i,J}}{dp_J}$ where $\mathbb{P}[J=j] = p_j$. Note that in the first setting we chose $p_j=\frac{1}{d}$, i.e. uniform sampling.
In the sparse setting, we chose $p_j$ to be nonzero only on the support of the points in question.
Informally, if $p_J$ is correlated with $z_{i,J}$, the estimator will have improved performance. Indeed, if $p_j\propto z_{i,j}$, then we would have that $X_i=\frac{z_{i,J}}{dp_J}= \frac{z_{i,J}\sum z_{i,J}}{dz_{i,J}} = \theta_i$, i.e. one sample gives us the exact value of $\theta_i$. However, computing this sampling distribution would be equivalent to solving the original problem. Thus the trade off arises between computing a good sampling distribution efficiently and solving the problem quickly given a good sampling distribution.
\subsection{Euclidean distances}
\label{sec:l2}
In this section we focus on $k$-NN graph construction (finding the $k$-NN of each point in the dataset) and consider the special case of $\ell_2$ distance.
We show how we can use the rotational invariance of the $\ell_2$ norm to improve the sub-Gaussian constants of our estimators.
Previously, we used (a bound on) the sub-Gaussian constant of our estimator $X_{i}$ as $\sigma_i$, allowing us to obtain confidence intervals.
As we will show, this sub-Gaussian constant can potentially be improved using the linear 
transformation $\mathcal{H}=HD$ where $H$ is the $d$ dimensional Hadamard matrix
and $D$ is the zero matrix with independent $\pm 1$ entries on the diagonal \cite{fjlt}.
This transformation can 
be computed in O$(nd \log d)$ time for 
$n$ points in $d$ dimensional space due to the recursive structure of the Hadamard matrix.
Due to the upfront cost, this only yields computational gains when amortized over the entire nearest neighbor graph construction.
We can then construct our new Monte Carlo box for this rotated data, $X_i^\texttt{R} = (x'_{0,J}-x'_{i,J})^2$ for $x'_{i} = \mathcal{H}x_i$, noting that applying $\mathcal{H}$
preserves pairwise Euclidean distances.
This rotation smooths out the coordinates with high probability, improving the concentration of our estimator.
This improvement is formalized in the following lemma, with proofs and further discussion in Appendix \ref{app:fjlt}.

\begin{lemma}\label{lem:rotatedEst}
The rotated Monte Carlo box $X_i^\texttt{R}$ satisfies $\|X_i^\texttt{R}\|_{\Psi_2} \le \frac{\|x_0-x_i\|^2_2\log \left(\frac{2n^2d}{\delta}\right)}{d}$ with probability at least $1-\delta$, compared to the original Monte Carlo box with $\|X_i \|_{\Psi_2} \le \frac{\|x_0-x_i\|^2_\infty}{2}$.
\end{lemma}
\noindent This constitutes an improvement by a factor of $\frac{d\|x_0-x_i\|^2_\infty}{2\|x_0-x_i\|^2_2\log \left(\frac{2n^2d}{\delta}\right)}$, potentially almost a factor of $d$. 
\section{Experimental Results} \label{sec:experiments}
In addition to the theoretical guarantees we prove for \texttt{BMO-NN} in this paper, 
we provide experimental results to demonstrate its practicality.
According to evaluations presented in the benchmark of
\cite{ann-benchmarks}, kGraph 
\cite{KGraph} and NGT \cite{iwasaki2018optimization} 
are two of the best performing practical $k$-NN
algorithms. 
We show our effective sample complexity gain (number of coordinate-wise distance computations made, as opposed to the base $nd$) in Figure \ref{fig:knn_compGain}.
\begin{figure*}[h!]
    \vspace{-.5cm}
    \centering
    \begin{subfigure}[b]{0.35\textwidth}
    \includegraphics[width=\textwidth, trim = 0 0 0 22, clip]{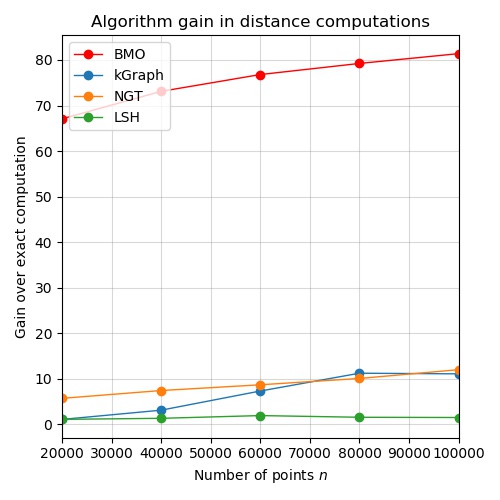}
    \vspace{-.7cm}
    \captionsetup{oneside,margin={.45cm,0cm}}
    \caption{}
    \end{subfigure}
    \begin{subfigure}[b]{0.35\textwidth}
    \includegraphics[width=\textwidth, trim = 0 0 0 22, clip]{Figures/numPulls_d.jpg}
    \vspace{-.7cm}
    \captionsetup{oneside,margin={.6cm,0cm}}
    \caption{}
    \end{subfigure}
    \vspace{-.1cm}
    \caption{Comparison of our \texttt{BMO-NN} algorithm
  with kGraph \cite{KGraph}, NGT \cite{iwasaki2018optimization}, and LSH \cite{andoni2015falconn_LSH}. Gain is measured versus number of operations for exact computation, with (a) varying $n$ and (b) varying $d$ (duplicate of Figure \ref{fig:BMONN_compGainIntro}).
  These plots are for exact $k$-NN querying only, index construction is not included. \texttt{BMO-NN} takes as input the target error probability $\delta= .01$ and achieves error probability less than 1\%, kGraph and LSH are tuned to have error probabilities of 1\%, and NGT with no tunable parameters has an error probability of $1-7\%$. Further details in Appendix \ref{app:details_of_experiments}.}
    \label{fig:knn_compGain}
    \vspace{-.6cm}
\end{figure*}

When run on the $100k$ Tiny ImageNet dataset with $k=5$, \texttt{BMO-NN} requires over 7x fewer coordinate-wise distance computations than these two state-of-the-art heuristic algorithms as shown in Figure \ref{fig:knn_compGain}.
Unlike these two, \texttt{BMO-NN} provides theoretical guarantees, while requiring over 50x fewer coordinate-wise distance computations than Falconn \cite{andoni2015falconn_LSH}, an approximation algorithm with guarantees.
Further, these algorithms all have expensive indices that need to be pre-computed (in stark contrast with \texttt{BMO-NN}) which were not factored into our computational gains plots.
Looking just at \texttt{BMO-NN}, we see in Figure \ref{fig:knn_compGain}(a) that the gain for \texttt{BMO-NN} changes very little as a function of the number of points $n$.
This is to be expected, as \texttt{BMO-NN}'s improvement comes from subsampling over the dimension $d$, as shown in Figure \ref{fig:knn_compGain}(b).
Here we see the dramatic, near linear gain of \texttt{BMO-NN} over exact computation as a function of $d$.
On the full dataset, \texttt{BMO-NN} provides an 80x improvement over exact computation.

Accuracy was held constant across operating points, with \texttt{BMO-NN} being run with a target error probability $\delta= .01$ and achieving an error probability of at most 1\% across simulations.
kGraph and LSH are tuned to have an error probability of 1\%, and NGT with no tunable parameters has an error probability of $1-7\%$.
Further details in Appendix \ref{app:details_of_experiments}.

One important question is whether the improvement afforded by our technique is due solely to the estimators (our Monte Carlo boxes), or if the adaptivity is important (Figure \ref{fig:block_dg}(b) versus (c)). From Figure \ref{fig:randSparseHist}(a) we can see that adaptivity is critical to the success of our algorithm, as uniform sampling these arms yields much worse performance, even at 80x the sample complexity of \texttt{BMO-NN}.

\begin{figure*}[h!]
  \centering
  \vspace{-.4cm}
  \includegraphics[width=.8\linewidth]{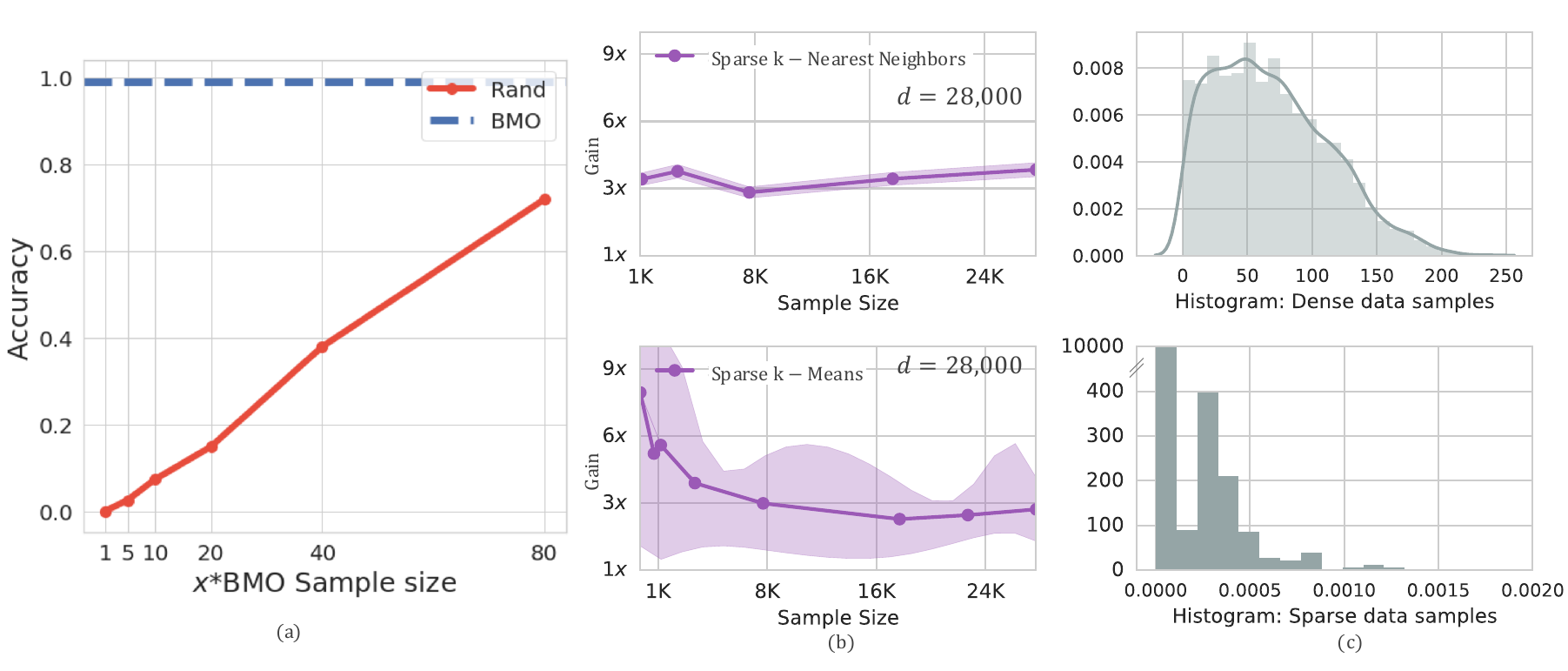}
  \vspace{-.4cm}
  \caption{(a) Non-adaptive Monte Carlo optimization has poor accuracy when computation is limited to $x$ times that used
  by \texttt{BMO-NN} on Tiny ImageNet.
  (b) Gain of \texttt{BMO-NN}
  on the sparse gene dataset of $10$\texttt{x}-genomics. Only computational gains against exact computation shown, as other algorithms do not take into account sparsity.
  (b) Histogram of coordinate-wise distances for randomly chosen pairs of points in the
  dense dataset (Tiny ImageNet) and sparse dataset (10x genomics).}\label{fig:randSparseHist}
  \vspace{-0.6cm}
\end{figure*}

As shown in Figure
\ref{fig:randSparseHist}(b) we
obtain a $3$x gain
over exact computation
in the number of coordinate-wise distance computations for $k$-nearest
neighbors on 
the sparse single cell gene expression dataset of \cite{10xdata}.
The baseline considered here takes
sparsity into account.
We also observe if we had directly used the estimator of Section
\ref{sec:knn} rather than the improved sparse Monte Carlo boxes in Section \ref{sec:sparsity},
we would not improve over exact computation.
Figure \ref{fig:randSparseHist}(c) shows that in real datasets our coordinate-wise distances have rapidly decaying tails, and so our sub-Gaussian assumption is not unreasonable.
We provide additional plots regarding random rotations in Figure \ref{fig:hadmard} in Appendix \ref{app:fjlt}.

\vspace{-.2cm}
\subsection{Numerical results for $k$-Means Clustering
}\label{sec:kmeans}

The canonical method for computing $k$-means,
Lloyd's algorithm \cite{Llo82}, starts with $k$ initial
centroids and then iteratively alternates between two steps --
\textit{the assignment step}: where each of the $n$ points in $\R^d$
is assigned to the centroid closest to it, and 
\textit{the update step}:
where the centroids are updated.  
In every iteration of a standard
implementation of $k$-means,
the assignment step takes $O(nkd)$
time per iteration while the update step
takes $O(nd)$ time per iteration.
We note that the assignment step of $k$ means
is nothing but finding the nearest neighbor of each of
the $n$ points 
among the $k$ centroids.  For each point, 
this can thus be posed as a nearest
neighbor problem with 
$k$ arms. If the distances are separable, then we can run \texttt{BMO-NN} to efficiently solve this problem.
As shown in 
Figure \ref{fig:kmeans_compGain} $(b)$, we obtain an improvement of 30-50x 
in terms of coordinate-wise distance computations over exact 
computation
on the Tiny ImageNet dataset 
with $k=100$ for $\ell_2$ distances.
In Figure
\ref{fig:randSparseHist}(b) we
show $3$x gain
over exact computation
on the single cell dataset of \cite{10xdata} using the sparse Monte Carlo boxes.
The results of Theorem \ref{thm:knn1} hold, as we are simply calling \texttt{BMO-NN} for 1-NN as a subroutine with $k$ arms instead of $n$.
Note that \texttt{BMO-NN} has gains in $d$ rather than in $n$ (the number of points), so we can still expect to see dramatic gains with $n=k$ cluster centers.

\begin{figure}[h]
  \begin{center}
%   \vspace{-.3cm}
    \includegraphics[width=\linewidth]{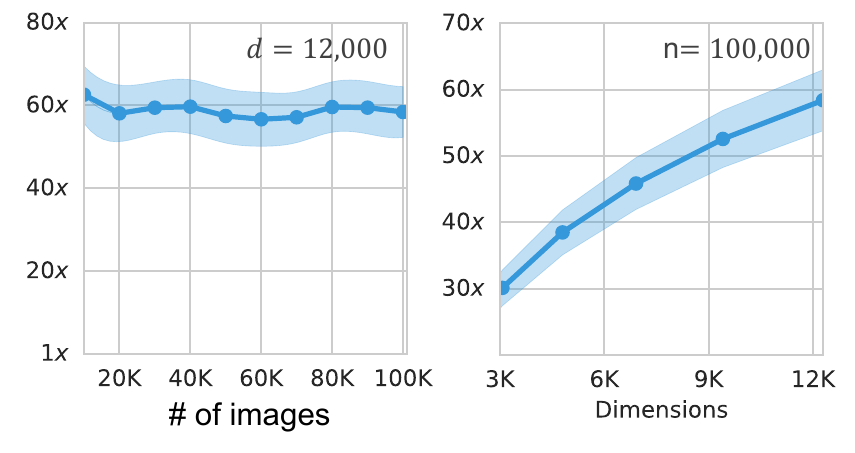}
    \vspace{-.8cm}
      \caption{Performance gain on the Tiny ImageNet dataset of BMO $k$-means over 
  exact computation
  in terms of number of computations needed ($k=100$). For this plot, BMO was constrained to have $> 99 \%$ accuracy.
  }\label{fig:kmeans_compGain}
\end{center}
\vspace{-.2cm}
\end{figure}

\vspace{-.2cm}
\subsection{Wall-clock time}

While the goal of this paper was not to optimize and implement a wall-clock efficient algorithm for $k$-NN, to show the potential real world practicality of \texttt{BMO-NN}, we implemented it in C++ to generate wall-clock results to show that adaptivity can indeed be efficient.
Comparing against exact computation and LSH, our two competitors with theoretical guarantees, we see that when run on the 100k Tiny ImageNet dataset for $k=5$ \texttt{BMO-NN} has a wall-clock time 5x faster than the LSH library of \cite{andoni2015falconn_LSH} as shown in Figure \ref{fig:knn_wallClockGain}. 
We used scikit-learn's nearest neighbors method as our exact computation baseline, which was the fastest off the shelf method we found \cite{scikit-learn}.
Even with all the compiler level optimization and batch efficiency in this method, our straightforward implementation of \texttt{BMO-NN} is able to outperform scikit-learn's optimized method by a factor of 1.5x on the Tiny ImageNet dataset.
To reduce the wall-clock time of our algorithm we made several modifications detailed in Appendix \ref{app:details_of_experiments}, including extending the initialization of \texttt{BMO UCB} to pull each arm 32 times, causing the running time to change very slowly with $d$.
These encouraging preliminary results show a promising direction of future work that optimizes the software performance to realize the 80x gain in number of computations, resulting in practical software libraries that can replace the state-of-the-art in some high dimensional applications.
\begin{figure*}[t]
\vspace{-.4cm}
    \centering
    \begin{subfigure}[b]{0.35\textwidth}
    \includegraphics[width=\textwidth, trim= 0 0 0 22, clip]{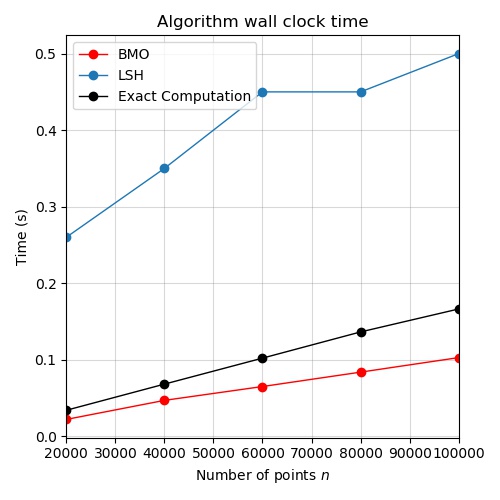}
    \captionsetup{oneside,margin={.45cm,0cm}}
    \vspace{-.7cm}
    \caption{}
    \end{subfigure}
    \begin{subfigure}[b]{0.35\textwidth}
    \includegraphics[width=\textwidth, trim= 0 0 0 22, clip]{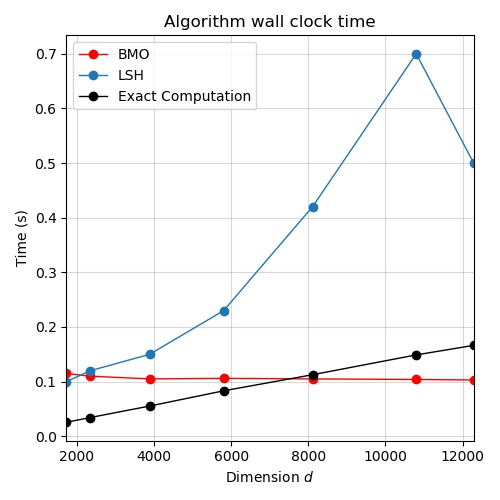}
    \captionsetup{oneside,margin={.6cm,0cm}}
    \vspace{-.7cm}
    \caption{}
    \end{subfigure}
    \caption{Comparison of our \texttt{BMO-NN} algorithm
  with kGraph \cite{KGraph}, NGT \cite{iwasaki2018optimization}, and LSH \cite{andoni2015falconn_LSH}. Gain is measured versus number of operations for exact computation, with (a) varying the number of points $n$ and (b) varying the dimension $d$.
  These plots are for $k$-NN querying only, index construction is not included. \texttt{BMO-NN} takes as input $\delta= .01$ and achieves error probability less than 1\%, kGraph and LSH are tuned to have error probability of 1\%, and NGT with no tunable parameters has error probability of $1-7$\%. Further details in Appendix \ref{app:details_of_experiments}.}
    \label{fig:knn_wallClockGain}
    \vspace{-.4cm}
\end{figure*}

\section{Discussions}\label{sec:disc}
In this paper we formalized the BMO technique and utilized it to solve the problem of $k$-nearest neighbors, achieving significant gains over state-of-the-art algorithms. 
Below we discuss lines of potential future work.

\subsection{Using bandit algorithms beyond UCB1} \label{sec:beyondUCB}

There is a large body of work on the pure exploration setting in multi-armed bandits for both the best-arm identification and top-$k$ identification problems.
There are numerous more sophisticated algorithms that one might wish to use instead of UCB1 in \texttt{BMO UCB}, like \cite{kalyanakrishnan2012pac,jamieson2014lil,simchowitz2017simulator}.
In this paper, our goal was to show that the BMO technique can be used to solve $k$-NN while requiring fewer coordinate-wise distance computations than the state-of-the-art, and so for simplicity we utilized UCB1.
Due to the computational overhead of adaptive algorithms in practice however, efficient implementations of these methods will necessarily pull several arms multiple times in each round in a batched fashion, and so such theoretical changes will not alter the running of the implemented algorithm.
For example, in every round of our implementation the top 32 arms are pulled 256 times each, as discussed in Appendix \ref{sec:implementation}. 
Several recent works have explicitly considered the batched setting \cite{jun2016top,jin2019efficient}, and may be of future interest.
% However, better theoretical bounds can perhaps be proved using the methods we describe here.

% There are however several steps of theoretical improvement that can be made beyond UCB1.
One potential way to improve the theoretical bounds in this paper is to use an algorithm like LUCB \cite{kalyanakrishnan2012pac} which is specifically designed for best-arm identification, as the analysis in this paper gives no nontrivial upper bound on the number of times we pull any of the top-$k$ arms.
The reason that UCB1 cannot be used for best-arm identification is that the second best arm may be pulled until its empirical mean minus its lower confidence bound (LCB) is only $\epsilon$ above the true mean of the best arm, in which case the best arm may need to be pulled until its confidence interval has width less than $\epsilon$.
Since we cannot give a lower bound on $\epsilon$ in this scenario, we cannot give a sample complexity bound for when the best arm’s confidence interval will separate from that of the second best arm.
However, this issue can be avoided by pulling the best arm and the arm with lowest LCB (ignoring the best arm) at each time step; this is the key idea in LUCB. Staying in best-arm identification literature, we can also attempt to employ an algorithm with order optimal sample complexity like lil'UCB \cite{jamieson2014lil}.
However, due to the novel stopping condition lil'UCB employs, it is less immediate how to incorporate the collapsing confidence intervals we have in this BMO setting, where if we sample an arm \texttt{MAX\_PULLS} times its confidence interval has width 0.

While LUCB is already an improvement that will eliminate the linear in $d$ additive term in front of the summation, there is still one more theoretical improvement that can be made.
Currently, and even with LUCB, we find the best arm, eliminate it, and find the best arm out of the remaining set, repeating until we have the top $k$.
However, if the top two arms have very close means, this is inefficient; we do not need to order the top $k$, we simply need to identify the set of top $k$ arms.
This motivates using a true top-$k$ identification algorithm like LUCB++, where the difficulty is measured in terms of the gaps between the top $k$ arms and the $k+1$-th arm, and the gaps between the $k$-th arm and the $k+1$-th to $n$-th arms \cite{simchowitz2017simulator}.
% \vspace{-.4cm}
\subsection{BMO extensions}

Modern datasets are often large in both the number of points and the number of dimensions. As mentioned in the introduction, most existing nearest neighbor approaches provide savings as the number of points scale, by exploiting constraints between distances such as the triangle inequality. On the other hand, our BMO-based approach provides savings as the number of dimensions scale, through sampling only a small subset of coordinates. Combining these approaches to obtain savings as {\em both} the number of points and the number of dimensions scale is an open problem of practical interest \cite{yang2021linear,mason2021nearest}.

In \cite{baharav2019Medoid} correlation in the underlying computational problem is exploited in the reduction to a bandit problem, yielding dramatic gains. This latter work suggests another avenue for further research on the BMO technique:  opening up the black box of the Monte Carlo samples. 
While the black box approach is a convenient way to convert the computational problem to a standard statistical multi-armed bandit one, it ignores any dependence that samples might have across boxes due to the underlying deterministic structure of the computational problem. Exploiting this structure can yield further improvement.

Another interesting avenue of future work is in developing dataset dependent lower bounds.
This was initially examined in \cite{lejeune2019adaptive}, but under very restrictive assumptions, namely that one could only interact with the data by sampling coordinates uniformly at random, and that each coordinate must have a value of $\pm 1/2$, leveraging  lower bounds from \cite{kaufmann2016complexity}.
Proving a lower bound in this more general case, allowing for correlation between arms, will require a more sophisticated analysis.

A final interesting observation is that we did not at any point need to generate independent unbiased estimates $X_i$ to run $\texttt{BMO UCB}$: we simply needed a sequence of unbiased estimators for each $\theta_i$ with
increasing accuracy 
$\{ \hat{\theta}_{i,\ell} \}_{\ell=1}^{\texttt{MAX\_PULLS}}$ with associated confidence intervals $\{ C_{i,\ell} \}_{\ell=1}^{\texttt{MAX\_PULLS}}$.
In this case, we need an analogue of Lemma \ref{lem:conf_intervals} to show that these confidence intervals hold with probability $1-\delta$. 
Then, we can prove an analogue of Theorem \ref{thm:knn1} stating that the number of samples needed for arm $i$ is upper bounded by $\min \left\{ \ell : C_{i,\ell} \le \frac{\Delta}{2}\right\}$.
In order for the BMO technique to yield an efficient solution in terms of number of computations needed, not just number of samples, we further require that updating 
$\hat{\theta}_{i,\ell}$ to $\hat{\theta}_{i,\ell+1}$ should be computationally cheap.
Finding $\theta_i$ that are not additive but but still permit such a sequence of estimators is an interesting line of future work.

% \vspace{-.2cm}
\section{Conclusion} \label{sec:conclusions}
Bandit-based Monte Carlo Optimization is a simple yet powerful technique. It converts a deterministic optimization problem into a statistical inference problem that can be efficiently solved via adaptive sampling. In this work, we highlight the broad applicability of the technique by applying it to the classical and important problem of nearest neighbor computation, and develop an algorithm that beats state-of-the-art methods in terms of both number of distance computations and wall-clock time, particularly in the high dimensional regime. 
Exploring the generality and limitations of BMO is a promising research direction that can unleash the full potential of this novel technique.

\bibliographystyle{IEEEtran}
\bibliography{ref}
% \clearpage
% !TEX root = main.tex
% !TEX spellcheck = en_US

\appendices

\section{Proofs for \texttt{BMO-NN}} \label{app:knn_sup}
\begin{proof}[Proof of Theorem \ref{thm:knn1}]
For notational simplicity, we assume that the $\theta_i$ are in sorted order, in that $\theta_1 \le \hdots \le \theta_n$ (equivalently $(i)=i$).
Let $T_i(t)$ be the number of times arm $i$ is pulled before iteration $t$ of \texttt{BMO UCB}. 
Additionally, let $\hat{\theta}_{i, T_i(t)}$ be the estimate of the 
mean of arm $i$ at the $t$-th iteration of the 
algorithm, and $2C_{i,T_i(t)}$ be the width of the
$(1-\delta')$ confidence interval of arm $i$ at iteration $t$
of the algorithm, for $\delta'=\frac{\delta}{nd}$. For any point $i$ which is not
one of the $k$ nearest neighbors
$\Delta_i^{(w)} \triangleq \theta_i-\theta_{w}$ measures how easy it is to determine that point $x_w$ is nearer than point $x_i$ to $x_0$. 
By this definition, 
$ \Delta_i^{(1)} \ge \Delta_i^{(2)} \ge \hdots \ge \Delta_i^{(k)}$.
Note that $\Delta_i^{(k)}= \Delta_i$ as defined in the Theorem.

We begin by analyzing the algorithm before 
it finds the first nearest neighbor of point $x_0$.
We observe
that if we choose to pull 
arm $i\neq 1$ at time $t$, then we have 
% \begin{align*}
$\hat{\theta}_{i, T_i(t)} - C_{i,T_i(t)} \leq \hat{\theta}_{1, T_{1}(t)} - C_{1,T_{1}(t)}$.  
% \end{align*}
For this to occur, at least one of the following $2n-1$ events must occur, for $\CE_1$, $\{\CE_2^{(i)}\}_{i= 2}^n, \{\CE_3^{(i)}\}_{i= 2}^n$:
\begin{align*}
& \mathcal{E}_1 = \left\{ \hat{\theta}_{1, T_{1}(t)} \geq \theta_{1}+C_{1,T_{1}(t)} \right\},\\
& \mathcal{E}_2^{(i)} = \left\{ \hat{\theta}_{i, T_i(t)} \leq \theta_i-C_{i,T_i(t)} \right\}, \\
& \mathcal{E}_3^{(i)} = \left\{ \Delta_i^{(1)} = \theta_i - \theta_{1} \leq 2 C_{i,T_i(t)} \right\}.
\end{align*}
This is because if all our estimators are within their confidence intervals, and the gap between the best arm and the rest is large compared to their confidence intervals, then our algorithm will pull the correct best arm.
Formally, note that if none of 
$\mathcal{E}_1, \mathcal{E}_2^{(i)}, \mathcal{E}_3^{(i)}$
occur,
we have that $\forall i \neq 1$
\begin{align*}
\hat{\theta}_{i, T_i(t)} - C_{i,T_i(t)} &\overset{(a)}{>} \theta_i - 2 C_{i,T_i(t)}\overset{(b)}{>}\theta_{1}\\
&\overset{(c)}{>} \hat{\theta}_{1, T_{1}(t)} - C_{1,T_{1}(t)},
\end{align*}
where $(a)$, $(b)$, and $(c)$  follow because $\mathcal{E}_2^{(i)}$, 
$\mathcal{E}_3^{(i)}$, and $\mathcal{E}_1$ do not hold
respectively.
By Lemma \ref{lem:conf_intervals} we have that our confidence intervals all hold simultaneously with probability at least $1-\delta$, that is
\begin{equation} \label{eq:sg_bound}
\P \left(\vert \theta_i - \hat{\theta}_{i, T_i(t)} \vert \leq C_{i,T_i(t)},~\forall~i\in[n],~\forall~t \right) \ge 1-\delta,
\end{equation}
and so $\mathcal{E}_1$ and $\{\mathcal{E}_2^{(i)}\}_{i \neq 1}$
do not occur during any iteration of \texttt{BMO-NN}
with probability at least $1-\delta$.
This also implies that 
with
probability at least $1- \delta$
the 
algorithm will have terminated when 
the events $\{\mathcal{E}_3^{(i)}\}$,
which are simple functions of the $T_i(t)$, stop occurring. 

Let $\zeta_i^{(1)}$ be the 
number of times the algorithm pulls arm $i$
before declaring the first nearest
neighbor of $x_0$ (line \ref{alg:line:addBest} of Algorithm \ref{alg:genUCB}).
From the previous discussion, we 
have that the algorithm will not pull arm $i$ more than $\zeta_i^{(1)}$ times
for the smallest value of $\zeta_i^{(1)}$ where the following holds.
\begin{align*}
&\sqrt{\frac{2 \sigma_i^2 \log \left( \frac{2nd}{\delta} \right)}{\zeta_i^{(1)}}} \le \frac{\Delta_i^{(1)}}{2} \text{ or } C_{i,\zeta_i^{(1)}} = 0\\
&\implies \zeta_i^{(1)}  \ge \frac{8 \sigma_i^2}{\left(\Delta_i^{(1)}\right)^2} \log 
\left( \frac{2nd}{\delta} \right) \text{ or }  \zeta_i^{(1)} \ge 2d.
\end{align*}
Hence, the total number of times we pull arm $i$ before declaring the first nearest neighbor of $x_0$ is at most 
$$\zeta_i^{(1)} = \left\lceil \frac{8 \sigma_i^2}{\left(\Delta_i^{(1)}\right)^2} \log \left( \frac{2nd}{\delta} \right)\right\rceil \wedge 2d,$$
at which point we have $\hat{\theta}_{i, T_i(t)} - C_{i,T_i(t)} > \hat{\theta}_{1, T_{1}(t)} - C_{ 1 ,T_{1}(t)}$, after which the algorithm only pulls arm 1.
The confidence interval $C_{1}$ will then keep shrinking, and since $C_{1}=0$ after $T_{1}=2d$ pulls, we will have $\forall i>1$ that $\hat{\theta}_{i, T_i(t)} - C_{i,T_i(t)} > \hat{\theta}_{1, T_{1}(t)} + C_{ 1 ,T_{1}(t)}$ within the next $d$ iterations, as arm 1 can only be pulled $2d$ times.
At this point the algorithm adds arm $1$ to the output set.
Therefore, the total number of coordinate-wise distances computed to obtain the 1-nearest neighbor of $x_0$ is at most $M_1 \le 2d+\sum_{i=2}^{n} \zeta_i^{(1)}$.

As \texttt{BMO-NN} progresses, let us similarly define
 $\zeta_i^{(w)}$ to be the 
number of times the algorithm pulls arm $i$
before declaring the $w$-th-nearest
neighbor for $w\in[k]$. 
By the same computation as above, we have that
\begin{align*}
\zeta_i^{(w)}  = \left\lceil \frac{8 \sigma_i^2}{\left(\Delta_i^{(w)}\right)^2} \log 
\left( \frac{2nd}{\delta} \right)\right\rceil \wedge 2d.
\end{align*}

As before, after pulling each arm $i> w$ at most $ \left\lceil \frac{8 \sigma_i^2}{\left(\Delta_i^{(w)}\right)^2} \log \left( \frac{2nd}{\delta} \right) \right\rceil  \wedge 2d$ times, the algorithm will then only pull arm $w$ until the upper confidence bound of $w$ separates from the lower confidence bound of all other arms (requiring at most $2d$ pulls), at which point arm $w$ will be added to the output set.

Note that we will pull each suboptimal arm $i$ as much as is necessary to identify each of the $k$ nearest neighbors, which for $i>k$ is dominated by the last term, i.e. $\max_{w \in [k]} \zeta_i^{(w)} = \zeta_i^{(k)}$.
Thus, the number of coordinate-wise distance computations needed to obtain all $k$ nearest neighbors is at most:
\begin{align*}
    M &\le 2kd+\sum_{i=k+1}^{n} \zeta_i^{(k)}\\
      &= 2kd+\sum_{i=k+1}^{n}\left( \left\lceil \frac{8 \sigma_i^2}{\Delta_i^2} \log \left( \frac{2nd}{\delta} \right) \right\rceil  \wedge 2d \right).
\end{align*}

Hence with probability at least $1- \delta$, the algorithm 
returns the $k$ nearest neighbors of $x_0$ with at most $M$ coordinate-wise distance computations, where as above we have that 
\begin{align*}
M
\le 2kd+\sum_{i = k+1}^{n} \left( \left\lceil \frac{8 \sigma_i^2}{\Delta_i^2} \log \left( \frac{2nd}{\delta} \right) \right\rceil  \wedge 2d \right).
\end{align*}

Note that this proved a slightly stronger statement: that \texttt{BMO-NN} returns the correct $k$-nearest neighbors \textit{in order of increasing $\theta_i$}.
Additionally, while the above steps implicitly assumed that $\theta_{1} < \hdots < \theta_{k} < \theta_{{k+1}} \le \hdots \le \theta_{{n}}$, our algorithm works as written for 
$\theta_{1} \le \hdots \le \theta_{k} < \theta_{{k+1}} \le \hdots \le \theta_{{n}}$ requiring only slight modifications to the proof (i.e. for $\Delta = 0$ letting $\frac{1}{\Delta^2} \wedge 2d = 2d$).
\end{proof}

\subsection{PAC scenario} \label{sec:knn_sup_PAC}

\begin{proof}[Proof of Theorem \ref{thm:knn_approx_sec}]
We make a similar argument here as in the proof for Theorem \ref{thm:knn1}, utilizing the same notation.
We begin by analyzing the algorithm before 
it finds the first approximate nearest neighbor of point $1$.
Consider the following events for all $i\in[n]$.
\begin{align*}
\mathcal{E}_1^{(i)} &= \left\{ |\hat{\theta}_{i, T_i(t)}-\theta_{i}| \geq C_{i,T_i(t)} \right\},\\
\mathcal{E}_2^{(i)} &= \left\{ 2C_{i,T_i(t)} \ge \max(\Delta_i,\epsilon) \right\}.
\end{align*}

We first show that the modified PAC \texttt{BMO-NN} outputs an $\epsilon$-best arm with probability at least $1-\delta$ and then use these events to bound the running time of the algorithm. 
We have that $\{\CE_1^{(i)} \}_{i=1}^n$ do not occur during any iteration with probability at least $1-\delta$, by Lemma \ref{lem:conf_intervals} on our $\delta' = \frac{\delta}{nd}$ confidence intervals.
Conditioned on these events not happening, we see that our algorithm will terminate in one of two scenarios.
In the first, the \texttt{BMO UCB} condition is satisfied (the upper confidence interval of the best arm separates from the lower confidence interval of the second best arm), in which case the algorithm outputs an arm $i$ such that for all $j\neq i$ we have
$ \theta_i 
\overset{(a)}{<}\hat{\theta}_{i, T_i(t)} + C_{i,T_i(t)} 
\overset{(b)}{<} \hat{\theta}_{j, T_j(t)} - C_{j, T_j(t)} 
\overset{(c)}{<} \theta_j,$
where a) is due to $\CE_1^{(i)}$, b) to the UCB termination condition, and c) to $\CE_1^{(j)}$.
Hence, in this case, we output the best arm.
In the other case, the algorithm selects to pull an arm $i$ with confidence interval $C_{i,T_i(t)}$ of width less than $\epsilon/2$ and terminates.
In this case, its output arm $i$ satisfies
$ \theta_i 
\overset{(a)}{<}\hat{\theta}_{i, T_i(t)} + C_{i,T_i(t)} 
\overset{(b)}{<}\hat{\theta}_{i, T_i(t)} - C_{i,T_i(t)} + \epsilon 
\overset{(c)}{\le} \hat{\theta}_{j, T_j(t)} - C_{j, T_j(t)} + \epsilon 
\overset{(d)}{<} \theta_j + \epsilon, $
where a) is due to $\CE_1^{(i)}$, b) to the stopping condition $C_{i,T_i(t)}<\epsilon/2$, c) is due to the fact that arm $i$ was to be pulled by the algorithm in the final step, and d) is due to $\CE_1^{(j)}$. Since this holds for all $j\neq i$, either $i=1$ in which case we ouput the best arm, otherwise we can set $j=1$ to see that $\theta_i < \theta_{1} + \epsilon$, and so $i$ is an $\epsilon$-best arm.
Hence, the algorithm successfully returns an $\epsilon$-best arm with probability at least $1-\delta$.

We now bound the number of pulls needed for each arm. We see that conditioned on the confidence intervals holding ($\{\CE_1^{(i)}\}, \{\CE_2^{(i)}\}$), an arm $i\neq 1$ can only be pulled while $\CE_2^{(i)}$ holds, as otherwise either $C_{i,T_i(t)} < \epsilon/2$ in which case the algorithm terminates if arm $i$ is selected to be pulled, or $C_{i,T_i(t)} < \Delta_i/2\le \Delta_i^{(1)}/2$, in which case
$ \hat{\theta}_{1, T_1(t)} - C_{1,T_{1}(t)} 
< \theta_{1} = \theta_i - \Delta_i^{(1)}
< \hat{\theta}_{i, T_i(t)} + C_{i,T_i(t)} - \Delta_i^{(1)} 
< \hat{\theta}_{i, T_i(t)} - C_{i,T_i(t)},$
showing that arm $i$ would not be selected to be pulled.
To compute how many pulls this requires, we see that $\CE_2^{(i)}$ does not hold when $C_{i,T_i(t)} < \max(\Delta_i^{(1)},\epsilon)/2$, which happens when
$$ \sqrt{\frac{2 \sigma_i^2 \log 2n^3d}{T_i(t)}} < \max(\Delta_i^{(1)},\epsilon)/2 \text{ or } T_i(t)=2d.$$
As in the proof of Theorem \ref{thm:knn1}, we 
have that the algorithm will not pull arm $i$ more than $\zeta_i^{(1)}$ times while finding the first approximate nearest neighbor of point $x_0$ where
\begin{align*}
\zeta_i^{(1)}  = \left\lceil\frac{8 \sigma_i^2 \log \left( \frac{2nd}{\delta} \right)}{(\Delta_i^{(1)} \vee \epsilon)^2}\right\rceil \wedge 2d.
\end{align*}
As the bandit algorithm progresses let us similarly define
 $\zeta_i^{(w)}$ to be the maximum
number of pulls of arm $i$ before declaring the $w$-th-nearest 
neighbor. 
By the same computation as above, we have
\begin{align*}
\zeta_i^{(w)}  = \left\lceil\frac{8 \sigma^2}{\left(\max(\Delta_i^{(w)},\epsilon)\right)^2} \log \left( \frac{2nd}{\delta} \right)\right\rceil \wedge 2d.
\end{align*}

Using the fact that $\zeta_i^{(w)}$
are increasing as a function of $w$,
the total number of distance computations
is
\begin{align*}
    M &= 2kd+\sum_{i=k+1}^{n} \zeta_i^{(k)}\\
      &= 2kd+\sum_{i=k+1}^{n} \hspace{-.1cm}\left( \left\lceil\frac{8 \sigma^2}{\left(\Delta_i \vee \epsilon \right)^2} \log \left( \frac{2nd}{\delta} \right)\right\rceil\wedge 2d \right).
\end{align*}
Hence our algorithm succeeds with probability at least $1-\delta$ in returning $\hat{i}_1,\hdots,\hat{i}_k$ such that $\theta_{\hat{i}_j}-\theta_{j} < \epsilon$ for $j\in[k]$ with a sample complexity bound as above.
\end{proof}

% \url{https://math.stackexchange.com/questions/1460412/concentration-of-maximum-of-gaussians}

\section{Complexity under Gaussian means} \label{app:knn_sup_normalproof}

In order to better understand what this gap dependence looks like for random data, we study the case where distances are drawn from a Gaussian distribution.
Since the bandit algorithm's performance depends only on the gaps between the arm means and not on the actual values themselves, the mean of the normal distribution we draw the distances from does not affect the analytical result.
However, in order for these Gaussian random variables to model our $k$-NN problem, distances must be nonnegative.
To this end, if the mean of the normal distribution is $\mu= \Omega\left(\sqrt{2\log n}\right)$ then all arms will have positive means with high probability.
Without loss of generality, the proof assumes that $\mu=0$, as the sample complexity depends only on the gaps.

\begin{proof}[Proof of Proposition \ref{cor:knn_normal}]
Denote the random number of coordinate-wise distance computations needed by \texttt{BMO-NN} as $M$, and let $(\cdot)$ be a permutation on $[n]$ such that $\theta_{(1)} \le \theta_{(2)} \le \hdots \le \theta_{(n)}$.
For clarity, we assume that $\sigma=1$ and $\mu=0$, but this does not affect the analysis.
We have from Theorem \ref{thm:knn1} that with probability at least $1 - \delta$, the correct set of $k$-NN will be returned with number of distance computations $M$ where
\begin{align*}
    &\mathbb{E}\left\{M\right\}
    \le \mathbb{E}\left\{2kd+\hspace{-.15cm}\sum_{i=k+1}^{n} \left( \left\lceil \frac{8\log \left( \frac{2nd}{\delta} \right) }{(\theta_{(i)}-\theta_{(k)})^2}  \right\rceil  \wedge 2d \right)\right\}\\
    &\le 2kd +n+ 16\sum_{i=1}^n \mathbb{E}\left\{ \frac{\log \left( \frac{nd}{\delta} \right)}{(\theta_i-\theta_{(k)})^2}  \wedge d \right\}. \numberthis
\end{align*}
We see that each term in this sum is identically distributed, and so we just need to bound one of them. We do this by dividing our range of $\theta_i$ into 2 regimes: $\theta_i - \theta_{(k)} < \tau_n $ and $\theta_i - \theta_{(k)} \ge  \tau_n$ for $\tau_n \triangleq c\frac{\log \log \left( \frac{nd}{\delta} \right)}{\sqrt{ \log \left( \frac{nd}{\delta} \right) }}$, where $c>0$ is some constant to be specified later.
Defining $G$ as the event where $\theta_i - \theta_{(k)} < \tau_n $, we show that the first regime occurs with low probability, and that our number of coordinate-wise distance computations needed is small in the second.
\begin{align*}
    \mathbb{E}\bigg\{ &\frac{\log \left( \frac{nd}{\delta} \right)}{(\theta_i-\theta_{(k)})^2}\wedge d \bigg\}\\
    &= \mathbb{E}\left\{ \frac{\log \left( \frac{nd}{\delta} \right)}{(\theta_i-\theta_{(k)})^2}\wedge d \ \middle\vert \ G\right\} \mathbb{P}\left( G \right) \\
    &\hspace{.1cm}+ \mathbb{E}\left\{ \frac{\log \left( \frac{nd}{\delta} \right)}{(\theta_i-\theta_{(k)})^2}\wedge d \ \middle\vert \ \bar{G}\right\} \mathbb{P}\left( \bar{G}\right) \numberthis\label{eq:gaussianSampleComplexity}
\end{align*}
We begin by analyzing the second overall case, where $\theta_i - \theta_{(k)} \ge \tau_n$ ($G$ did not occur). We bound the probability trivially as 
$\mathbb{P}\left(\bar{G}\right) \le 1$. We see that expectation is well behaved, with
\begin{equation} \label{eq:largeDeltaBound}
\hspace{-1cm}\mathbb{E}\left\{ \frac{\log \left( \frac{nd}{\delta} \right)}{(\theta_i-\theta_{(k)})^2}\wedge d \ \middle\vert \ \bar{G}\right\} \le \left(\frac{\log \left( \frac{nd}{\delta} \right)}{c\log \log \left( \frac{nd}{\delta} \right)} \right)^2\hspace{-.2cm},\hspace{-.5cm}    
\end{equation}
which concludes our analysis of the second term.
We now examine the first term, when $\theta_i - \theta_{(k)} < \tau_n$ ($G$ did occur), noting that

\begin{equation} \label{eq:smallGapVal}
    \mathbb{E}\left\{ \frac{\log \left( \frac{nd}{\delta} \right)}{(\theta_i-\theta_{(k)})^2}\wedge d \ \middle\vert  \ G \right\} \le d.
\end{equation}
\noindent The rest of the proof is simply bounding the probability that the gap $\theta_i - \theta_{(k)}$ is small.
Here we use the assumption that $\mu=0$, that is $\E \{\theta_i\} = 0$, to avoid centering the $\theta_i$ each time. This is without loss of generality. 
We see that in order for the gap to be small either $\theta_i$ must have been small, or $\theta_{(k)}$ must have been large. Defining $\gamma_n= \sqrt{2 \log n - K\log \log n}$, we have that
\begin{align*}
    \mathbb{P}(G)
    \le &\mathbb{P}\left(\left\{\theta_i + \gamma_n < \tau_n\right\} \cup \left\{\theta_{(k)} > -\gamma_n \right\}\right)\\
    \le &\mathbb{P}\left(\theta_i+\gamma_n < \tau_n\right) + \mathbb{P}\left(\theta_{(k)} > -\gamma_n \right), \numberthis \label{eq:smallGapCases}
\end{align*}
where $K>0$ is a constant to be specified later.
We begin by bounding the first term in \eqref{eq:smallGapCases} as
\begin{align*}
    \mathbb{P}\bigg(\theta_i + \gamma_n< \tau_n\bigg)
    &= \mathbb{P}\left(\theta_i < -\gamma_n + \tau_n\right)\\
    &\underset{(a)}{\le} e^{- \left( -\gamma_n + \tau_n\right)^2/2}\\
    &= e^{-\left(\gamma_n^2  -2\tau_n\gamma_n + \tau_n^{2}\right)/2}\\
    &\le e^{-\log n  +(K/2 + c \sqrt{2})\log \log \left( \frac{nd}{\delta} \right) }\\
    &= \frac{\left(\log \left( \frac{nd}{\delta} \right)\right)^{K/2 + c \sqrt{2}}}{n}, \numberthis \label{eq:orderStat1}
\end{align*}
where (a) holds when $\gamma_n > \tau_n$, which is true for all $n>1$ as we will have $K<4$ and $c<1$.
To show that $\mathbb{P}\left(\theta_{(k)} > -\gamma_n \right) = O\left(\frac{1}{n}\right)$ when $K> 3$ requires a more sophisticated order statistics analysis:
\begin{align*}
    \mathbb{P}&\bigg(\theta_{(k)} > -\gamma_n \bigg)\\
    &= \sum_{i=0}^{k-1}{n \choose i} \left(\mathbb{P}\left(\theta_1 > -\gamma_n\right) \right)^{n-i} \left(\mathbb{P}\left(\theta_1 < -\gamma_n\right) \right)^i\\
    &\le \left(\sum_{i=0}^{k-1} n^i \right)\left(\mathbb{P}\left(\theta_1 > -\gamma_n\right) \right)^{n-k}\\
    &\le 2n^{k-1} \left(\mathbb{P}\left(\theta_1 > -\gamma_n\right) \right)^{n-k}. \numberthis
\end{align*}
In the last line we use $n\ge 2$ to simplify the sum.
We can now perform a standard order statistics computation, which for completeness we include below.
This uses a bound on the Gaussian CCDF $Q$, the fact that $k<n/2$, and a lot of algebra to show that:
\begin{align*}
\bigg(\mathbb{P}&\left(\theta_1 > -\gamma_n\right) \bigg)^{n-k}\\
    &\le \left(\mathbb{P}\left(\theta_1 > -\gamma_n \right)\right)^{n/2}
    = \left(1 - Q\left(\gamma_n \right)\right)^{n/2}\\
    % &\le \left(1 - \frac{\sqrt{2 \log n - K\log \log n}}{1 + \left(\sqrt{2 \log n - K\log \log n}\right)^2} e^{-\left(\sqrt{2 \log n - K\log \log n}\right)^2/2}\right)^{n/2}\\
    &\le \left(1 - \frac{\gamma_n}{1 + \gamma_n^2} e^{-\gamma_n^2/2}\right)^{n/2}\\
    &\le \left(1 - \frac{1}{1 + \gamma_n} e^{-\log n + \frac{K}{2}\log \log n}\right)^{n/2}\\
    &\le \exp\left( - \frac{n/2}{1 + \gamma_n} e^{-\log n + \frac{K}{2}\log \log n}\right)\\
    &\le \exp\left( - \frac{\left(\log n \right)^{K/2}}{2 + 2\gamma_n}\right)\\
    &\le \exp\left( - \frac{\left(\log n \right)^{K/2}}{2 + 2\sqrt{2 \log n}}\right)\\
    &\le \exp\left( -\left(\log n \right)^{(K-1)/2}/6\right)\\
    % &= O\left(\exp\left( - \left(\log n \right)^{(K-1)/2}\right)\right)\\
    &= n^{\left( - \frac{1}{6}\left(\log n \right)^{(K-3)/2}\right)}, \numberthis
\end{align*}
where we used that $n\ge 2$ and so $2+2\sqrt{2\log n}\le 6\sqrt{\log n}$. Thus for $k \le \frac{1}{6}\left(\log n \right)^{(K-3)/2}$ we have
\begin{align*}
    \mathbb{P}(\theta_{(k)} > -\gamma_n &)
    \le 2n^{k-1} \left(\mathbb{P}\left(\theta_i > -\gamma_n\right) \right)^{n-k}\\
    &\le 2n^{k-1} n^{\left( - \frac{1}{6}\left(\log n \right)^{(K-3)/2}\right)}
    \le \frac{2}{n} \numberthis \label{eq:orderStat2}
\end{align*}
Hence, plugging \eqref{eq:orderStat1} and \eqref{eq:orderStat2} into \eqref{eq:smallGapCases}, we have that:
\begin{align*}
    \mathbb{P}\left(\theta_i - \theta_{(k)} \hspace{-.05cm}< \hspace{-.05cm}\tau_n\right)
    &\le \mathbb{P}\left(\theta_i +\gamma_n \hspace{-.05cm}<\hspace{-.05cm} \tau_n\right) + \mathbb{P}\left(\theta_{(k)} \hspace{-.05cm}>\hspace{-.05cm} -\gamma_n \right) \\ 
    &\le \frac{\left(\log \left( \frac{nd}{\delta} \right)\right)^{K/2 + c \sqrt{2}}}{n} + \frac{2}{n}\\
    &\le \frac{2\left(\log \left( \frac{nd}{\delta} \right)\right)^{K/2 + c \sqrt{2}}}{n}. \numberthis \label{eq:smallGapProb}
\end{align*}
Plugging \eqref{eq:smallGapProb}, \eqref{eq:smallGapVal}, and \eqref{eq:largeDeltaBound} into \eqref{eq:gaussianSampleComplexity}, we bound the expected sample complexity for a random arm
using $G$ as the event where $\theta_i - \theta_{(k)} < \tau_n $, yielding
\begin{align*}
    \mathbb{E}\bigg\{ &\frac{\log \left( \frac{nd}{\delta} \right)}{(\theta_i-\theta_{(k)})^2}\wedge d \bigg\}\\
    &= \mathbb{E}\left\{ \frac{\log \left( \frac{nd}{\delta} \right)}{(\theta_i-\theta_{(k)})^2}\wedge d \ \middle\vert \ G\right\} \mathbb{P}\left( G \right) \\
    &\hspace{.1cm}+ \mathbb{E}\left\{ \frac{\log \left( \frac{nd}{\delta} \right)}{(\theta_i-\theta_{(k)})^2}\wedge d \ \middle\vert \ \bar{G}\right\} \mathbb{P}\left( \bar{G}\right) \\
    &\le d \cdot \frac{2\left(\log \left( \frac{nd}{\delta} \right)\right)^{K/2 + c \sqrt{2}}}{n}
    +\left(\frac{\log \left( \frac{nd}{\delta} \right)}{c\log \log \left( \frac{nd}{\delta} \right)} \right)^2\\
    &= O\left((1+d/n)\frac{\log^2 \left( \frac{nd}{\delta} \right)}{\log^2 \log \left( \frac{nd}{\delta} \right)} \right). \numberthis
\end{align*}
Where in the last line we
set $K=4-3c<4-2\sqrt{2}c$ for constant $0<c<\sqrt{2}/4$ to get $K/2+c\sqrt{2} < 2$.
This means we can accommodate $k \le \frac{1}{6}\left(\log n \right)^{(K-3)/2}= \frac{1}{6}\left(\log n \right)^{1/2-3c/2}$.
Thus, our expected total number of coordinate-wise distance computations is
\begin{align*}
\mathbb{E} \{M\}
&\le 
2kd + n + 16\sum_{i=1}^n \mathbb{E}\left\{ \frac{\log \left( \frac{nd}{\delta} \right)}{(\theta_i-\theta_{(k)})^2}  \wedge d \right\}\\
&= O\left( \left(n+d\right)\frac{\log^2 \left( \frac{nd}{\delta} \right)}{\log^2 \log \left( \frac{nd}{\delta} \right)} \right).
\end{align*}

We remark that one could have chosen the analysis threshold $\tau_n$ as $\frac{c\log \log n}{\sqrt{ \log n}}$ instead of $\tau_n= \frac{c\log \log \left( \frac{nd}{\delta} \right)}{\sqrt{ \log \left( \frac{nd}{\delta} \right)}}$ to achieve a slightly tighter bound, but there will regardless be a $\log \left( \frac{nd}{\delta} \right)$ factor in the final expression, and in the regime where $d=\text{poly}(n)$ and $\delta = \text{poly}\left(\frac{1}{n}\right)$ the bound will be equivalent.
\end{proof}

\section{Improved Monte Carlo Boxes} \label{app:MCboxes}
\subsection{Details about the sparse estimator}\label{app:sparsity}

Let the set of non-zero entries of $x_0$ and $x_i$ be $S_0 \subseteq [d]$ and $S_i \subseteq [d]$ respectively, with $|S_0| = n_0$ and $|S_i| = n_i$, and $N=n_0+n_i$. Using $\ell_1$ distance for concreteness, recall that our sparse estimator $X_i^\texttt{S}$ was defined as:
\begin{equation*}
X_i^\texttt{S}\hspace{-.05cm}= \hspace{-.08cm}
\begin{cases}
\hspace{-.05cm}\frac{N }{2d} \left|x_{0,t^{(0)}}\hspace{-.01cm}-\hspace{-.01cm}x_{i,t^{(0)}}\hspace{-.03cm}\right| (1+ \mathds{1}\{t^{(0)} \hspace{-.05cm} \not\in S_i\}) & \text{w.p. } \frac{n_0}{N}\\
\hspace{-.05cm}\frac{N}{2d}\left|x_{0,t^{(i)}}-x_{i,t^{(i)}}\right|
    (1+ \mathds{1}\{t^{(i)} \hspace{-.05cm} \not\in S_0\}) & \text{w.p. } \frac{n_i}{N}\\
\end{cases}    
\end{equation*}
This can intuitively be seen as flipping a biased coin to see which support we draw a coordinate from, and then multiplying by 2 if this coordinate does not appear in the other point's support (to offset the double counting).
We see that $X_i^\texttt{S}$ is unbiased, as we can analytically rewrite it as 
$$X_i^\texttt{S}=
\begin{cases}
\frac{N }{d} \left|x_{0,j}-x_{i,j}\right| &\text{w.p. } \frac{1}{N} \text{ for }j \in S_0 \Delta S_i\\
\frac{N }{2d} \left|x_{0,j}-x_{i,j}\right| &\text{w.p. } \frac{2}{N} \text{ for }j \in S_0 \cap S_i
\end{cases}
$$
where $S_0\Delta S_i=(S_0 \cup S_i) \setminus (S_0 \cap S_i)$ is the symmetric difference of $S_0$ and $S_i$.
% More concretely, 
% \begin{align*}
%     &\E X_i^\texttt{S} \\
%     &= \frac{1}{n_0+n_i} \sum_{j\in S_0}\frac{n_0+n_i }{2d} \left|x_{0,j}-x_{i,j}\right| (1+ \mathds{1}\{j \not\in S_i\})\\
%     &\hspace{.2cm}+ \frac{1}{n_0+n_i} \sum_{j \in S_i} \frac{n_0+n_i}{2d}\left|x_{0,j}-x_{i,j}\right|
%     (1+ \mathds{1}\{j \not\in S_0\})\\
%     % &=  \frac{1}{2d} \sum_{j\in S_0} \left|x_{0,j}-x_{i,j}\right| (1+ \mathds{1}\{j \not\in S_i\})
%     % + \frac{1}{2d} \sum_{j \in S_i} \left|x_{0,j}-x_{i,j}\right|(1+ \mathds{1}\{j \not\in S_0\})\\
%     &=  \frac{1}{2d} \sum_{j\in S_0} \left|x_{0,j}-x_{i,j}\right| 
%         + \frac{1}{2d} \sum_{j\in S_0 \setminus S_i} \left|x_{0,j}-x_{i,j}\right|\\
%     &\hspace{.2cm}
%         + \frac{1}{2d} \sum_{j\in S_i} \left|x_{0,j}-x_{i,j}\right|
%         + \frac{1}{2d} \sum_{j\in S_i \setminus S_0} \left|x_{0,j}-x_{i,j}\right|\\
%     &= \frac{1}{d}\sum_{j \in S_0 \cup S_i} \left|x_{0,j}-x_{i,j}\right| = \frac{1}{d}\|x_0-x_i\|_1 \numberthis
% \end{align*}
Hence $X_i^\texttt{S}$ is unbiased.
Next, we provide the proof of Lemma \ref{lem:sparseEst}, showing the improvement afforded by utilizing the sparsity.

\begin{proof}[Proof of Lemma \ref{lem:sparseEst}, Sparse Estimator]
Initially we have that our $X_i$ is bounded in the interval $[0,\max_{j\in S} |x_{0,j}-x_{i,j}| ]$.
Now, our new sparse estimator $X_i^\texttt{S}$ is bounded in the interval $[\min_{j\in S}\frac{n_0+n_i}{2d}|x_{0,j}-x_{i,j}|,  \max_{j\in S}\frac{n_0+n_i}{d}|x_{0,j}-x_{i,j}| ]$, or more coarsely, $[0, \frac{n_0+n_i}{d}\max_{j\in S}|x_{0,j}-x_{i,j}|$.

By Hoeffding's Lemma, if a random variable is bounded between $[a,b]$ then it is sub-Gaussian with parameter $(b-a)/2$.
Our initial estimator then has $\sg{X_i}\le \frac{1}{2}\max_{j\in S} |x_{0,j}-x_{i,j}|$, and our new estimator has $\sg{X_i^\texttt{S}} \le \frac{n_0+n_i}{d}\max_{j\in S} |x_{0,j}-x_{i,j}|$.
This is a factor of $\left(\frac{d}{2(n_0+n_i)}\right)$ improvement.
\end{proof}

Generating this unbiased estimate $X_i^\texttt{S}$ 
thus requires only two 
non-trivial operations; $1)$ sampling
from the non-zero coordinates of a sparse vector, and $2)$
checking if a coordinate is non-zero in a sparse vector.
These operations can generally be performed in $O(1)$ time, as in many settings sparse vectors are stored in
a data-structure with two sub-structures:
a vector $v_{\text{data}}$ storing the non-zero entries
and
a dictionary (or unordered map)
which maps the coordinate in the original
vector to the coordinate in  $v_{\text{data}}$. 
An entry with a value of $0$ is not in the dictionary. 
We can check the membership of a key in a 
dictionary as well as find the value of the key
if it is present in the dictionary
in $O(1)$ time. Additionally, we can
generate a random key in the dictionary in $O(1)$ time. 
Assuming that our sparse vectors are stored in this manner gives the desired $O(1)$ update time for our estimators.

\subsection{Improving sub-Gaussian constant via random rotations for Euclidean distance} \label{app:fjlt}
Consider the problem of computing nearest neighbors under
Euclidean distance (which is the same as k-NN 
under squared Euclidean distance)
as we did in Section \ref{sec:l2}. 
The estimator we proposed was $X_i =(x_{1,t}- x_{i,t})^2$ for $t \sim \text{Unif}([d]).$
Recall that $\theta_i = \frac{1}{d}\|x_1-x_i\|_2^2$ and that we denote 
the sub-Gaussian parameter of $X_i$ as $\sigma_i$. 
To gain some intuition regarding $\sigma_i$, note that if the 
distance between all coordinates of 
$x_1$ and $x_i$ are the same then $\sigma_i = 0$. 
On the other hand, if $x_1$ and $x_i$ agree on 
all coordinates but one, then the sub-Gaussian
parameter $\sigma_i \le \frac{1}{2}\|x_1-x_i \|_2^2 = \frac{1}{2}d\theta_i$. This coarse bound is reasonably accurate, as the sub-Gaussian parameter of a Bernoulli random variable with heads parameter $p=\frac{1}{d}$ is $\frac{d-2}{2d\log(d-1)} \ge \frac{1}{4\log(d)}$ for $d>10$, so $\|X_i \|_{\Psi_2} \ge \frac{1}{4\log(d)} \|x_1-x_i\|_2^2 = \frac{d}{4\log(d)} \theta_i$ \cite{subg_binaryRV}.
If the squared difference in each 
coordinate were bounded between $0$ and $\gamma$, then
$X_i$ would be 
$\gamma/2$-sub-Gaussian by Hoeffding's Lemma. 
We show how a primitive called random rotations \cite{fjlt} can improve this sub-Gaussian parameter, achieving this last result with $\gamma \approx \theta_i\sqrt{ \frac{\log \left( \frac{nd}{\delta}\right)}{d}}$.

Let $D\in \R^{d \times d}$ be a diagonal 
matrix where each diagonal entry 
is independently $\pm1$ with equal probability i.e, 
\begin{equation}
    D = \text{diag}(Y_1,\hdots,Y_d) \hspace{.5cm}, \hspace{.5cm} Y_i \overset{iid}{\sim} \text{Unif}(\{\pm 1\}) \label{eqn:diagonal_matrix}
\end{equation}
A $d$ dimensional Hadamard matrix $H_d$ is
recursively defined as follows (we assume $d$ is a power of 2, if not one can achieve the same results up to constants via zero padding):
\begin{align}
    H_2 \hspace{-.05cm}&= \hspace{-.05cm}\frac{1}{\sqrt{2}}\hspace{-.05cm}\begin{bmatrix}
    1 \hspace{-.1cm}& 1 \\
    1 \hspace{-.1cm}& -1
  \end{bmatrix},
    \ H_{2^{k+1}} \hspace{-.05cm}= \hspace{-.05cm}\frac{1}{\sqrt{2}}\hspace{-.05cm}\begin{bmatrix}
    H_{2^{k}}\hspace{-.1cm} & H_{2^{k}} \\
    H_{2^{k}}\hspace{-.1cm} & -H_{2^{k}}
  \end{bmatrix}\label{eqn:hadamard_matrix}
\end{align}
A random rotation consists of preprocessing each point $x_i$ by rotating it with the random rotation matrix $\CH = HD$ to obtain $x'_i =  HDx_i$. The $\ell_2$ distance is invariant under rotation and so we have $\|x'_i - x'_j\|_2 = \|x_i - x_j\|_2$. 
An unstructured rotation in $d$ dimension requires $O(d^2)$ time to compute per point. However, due to the recursive structure of the Hadamard matrix $H$, the rotation $HD$ can be applied in $O(d\log d)$ time. 
The corresponding estimator for the post processed points $\{x'_i\}_
{i=1}^{n}$ is
\begin{align}\label{eq:knn1_repeat_rotate}
     X_i^\texttt{R} &= 
     (x'_{1,J}- x'_{i,J})^2, \quad J \sim \text{Unif}([d]).
\end{align}

We now give a helpful lemma, the proof of which is almost identical to that of Lemma 1 from \cite{fjlt}.
\begin{lemma}  \label{thm:hadamard_l_infinity_mt}
For $x'_{i} = \mathcal{H}x_i$, with probability at least $1-\delta$ we have that for all $i,j\in[n]$ simultaneously
\begin{equation}
\|x'_i - x'_j\|_{\infty} \leq \|x_i - x_j\|_2\sqrt{\frac{2\log \left(\frac{2n^2d}{\delta}\right)}{d}}.    
\end{equation}
\end{lemma}
\begin{proof} \label{proof:rotation_ub_subg}
Define the random variable $u = HD(x_1-x_2) = \{u_1, u_2, \cdots, u_d\}$. Note that $u_1$ is of the form $\sum_{i=1}^da_i(x_{1,i}-x_{2,i})$ where each $a_i \in \{d^{-1/2},-d^{-1/2}\}$ is chosen independently and uniformly at random. Using a Chernoff-type argument we have by symmetry that for any $t>0$
\begin{align*}
    \mathbb{P}\left(|u_1|>s\right) &= 2\mathbb{P}\left(e^{tu_1}\geq e^{ts}\right)  \leq 2\mathbb{E}\left[e^{tu_1}\right]/e^{ts}\\
    &\leq 2e^{t^2\| x_1 - x_2\|^2_2/2d-ts}.
\end{align*}
Substituting $t=\frac{sd}{\|x_1 - x_2\|^2_2}$ we obtain $\mathbb{P}\left(|u_1|>s\right)\leq 2e^{-\frac{s^2d}{2\| x_1 - x_2\|^2_2}}$, and setting $s=\|x_1 - x_2\|_2\sqrt{\frac{2\log \left(\frac{2n^2d}{ \delta}\right)}{d}}$ yields
\begin{align*}
    \mathbb{P}\left(|u_1|>\|x_1 - x_2\|_2\sqrt{\frac{2\log \left(\frac{2n^2d}{\delta}\right)}{d}}\right)\leq \frac{\delta}{n^2d}
\end{align*}
Taking a union bound over $d$ coordinates gives us
\begin{align*}
    \mathbb{P}\left(\|u\|_{\infty}>\|x_1 - x_2\|_2\sqrt{\frac{2\log \left(\frac{2n^2d}{\delta}\right)}{d}}\right)\leq \frac{\delta}{n^2}.
\end{align*}
Taking a further union bound over all pairs of points $(x_i,x_j)$ gives us the desired result.
\end{proof}

From this we are able to prove Lemma \ref{lem:rotatedEst}.
\begin{proof}[Proof of Lemma \ref{lem:rotatedEst}, Rotated Estimator]
We have by Hoeffding's Lemma that $\sg{X_i} \le \frac{1}{2}\|x_1 - x_i \|_\infty^2$.
Similarly, for the rotated estimator we have that with probability at least $1-\delta$
\begin{align*}
\sg{X_i^\texttt{R}} 
&\le \frac{1}{2}\|x_1' - x_i' \|_\infty^2
\underset{(a)}{\le} \|x_1 - x_i\|_2^2 \frac{\log \left(\frac{2n^2d}{\delta}\right)}{d}\\
&= \theta_i \log \left(\frac{2n^2d}{\delta}\right),
\end{align*}
where (a) comes from Lemma \ref{thm:hadamard_l_infinity_mt}, which fails with probability at most $\delta$.
\end{proof}

\subsubsection{Additional numerical results for random rotations} \label{sec:rotation_additional_simulations}
To empirically show the effects of this procedure, we applied random rotations as per Lemma
\ref{thm:hadamard_l_infinity_mt} on several images from the Tiny ImageNet dataset.
In Figure \ref{fig:hadmard}, the top row is the empirical distribution of the coordinate-wise squared distance between $4$ pairs of images represented by  $\big\{(x_{1}, x_{2}), (x_{3}, x_{4}), (x_{5}, x_{6}), (x_{7}, x_{8}) \big\} $.
\begin{figure}[h]
  \centering
  \includegraphics[width=.9\linewidth,trim= 0 0 0 42, clip]{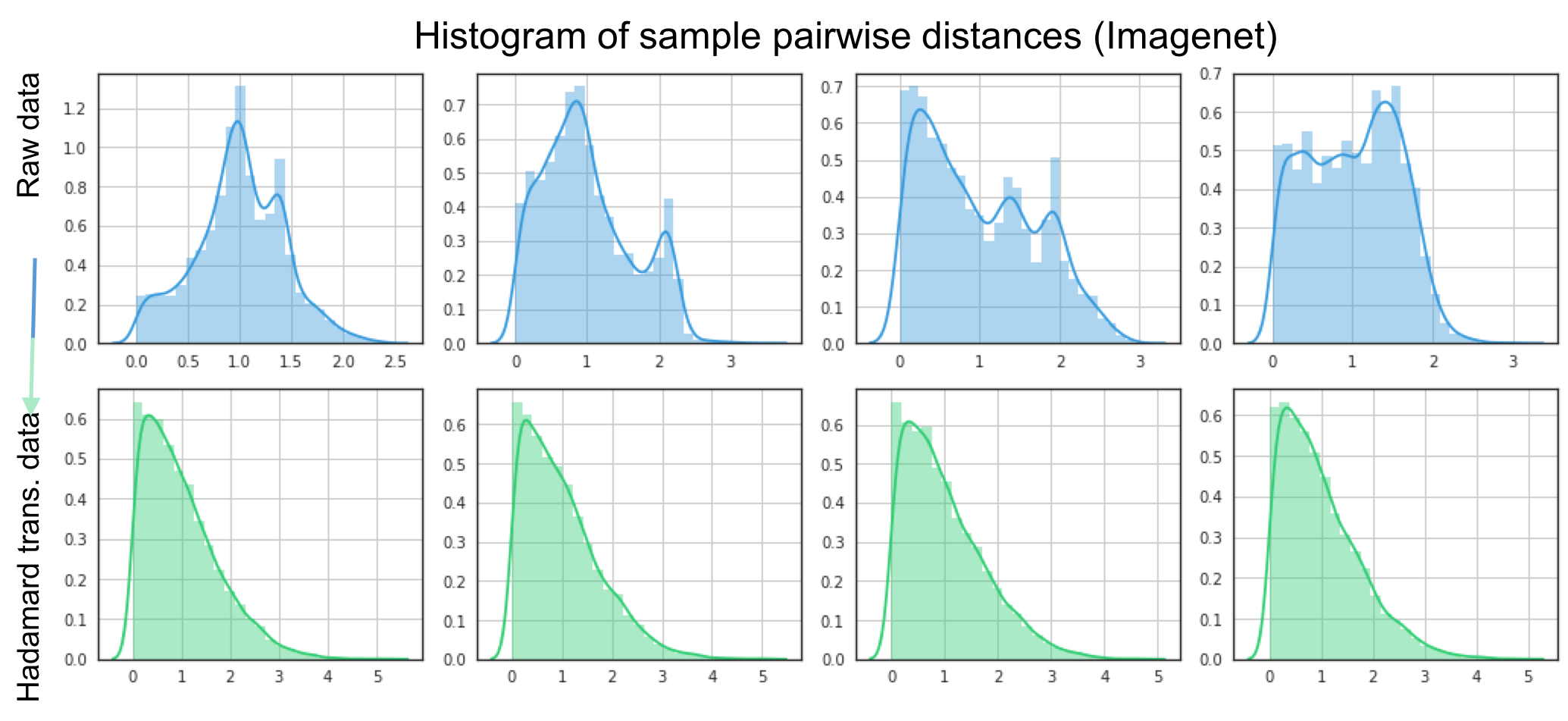}
  \caption{Histogram of pairwise distances from ImageNet. Top row: histogram of $(x_i[l] - x_j[l])^2 \; \forall \; l\in [d]$. Bottom row: histogram of $(x'_i[l] - x'_j[l])^2 \; \forall \; l\in [d]$, where $x'_i = HDx_i.$ }\label{fig:hadmard}
\end{figure}
Each of these vectors is rotated as $x'_i = HDx_i$ (equations \eqref{eqn:diagonal_matrix}, \eqref{eqn:hadamard_matrix}). The bottom row is the coordinate-wise squared distance between the same $4$ pairs of images represented by $\big\{(x'_{1}, x'_{2}), (x'_{3}, x'_{4}), (x'_{5}, x'_{6}), (x'_{7}, x'_{8}) \big\} $ instead. We can see that the histograms in the bottom row have lighter tails compared to their counterparts in the top row.

\section{Details of experiments} \label{app:details_of_experiments}
\subsection{Software} \label{sec:implementation}
We implemented \texttt{BMO-NN} in C++ in about 1500 lines of code. The results and figures of the paper can be
reproduced from the publicly available code.
We now briefly describe the design architecture of our implementation and important optimizations needed to outperform the state-of-art algorithms in both FLOPs and wall-clock time.
\paragraph{Architecture} 
The \texttt{BMO UCB} routine (as described in Algorithm \ref{alg:genUCB}) takes in a set of arms and outputs the top $k$ arms.
\texttt{BMO-NN} takes the dataset as an input and defines the `arms' with `pull' and `update' functions and then calls the \texttt{BMO UCB} routine. The \texttt{BMO UCB} routine is the most complex and heavy part, and so we optimized \texttt{BMO UCB} in terms of computation time and read-write to the disk. On the other hand, defining arms and their functions is light weight, and thus it easy to implement an adaptive version of a new problem in the BMO framework. 

\paragraph{Changes from written algorithm} \texttt{BMO UCB} as stated pulls each arm once, then proceeds by pulling one arm in each iteration once.
To improve our wall-clock time in practice, we begin by pulling each arm 32 times. Then, in each iteration we pick the 32 arms with lowest mean-LCB and pull each one of them 256 times.
Theoretically this can only increase the number of samples needed by a constant factor, and in practice performs well.
Additionally, we sample the coordinates with replacement for the sake of simplicity, where sampling without replacement would reduce the number of samples needed but increase the computational overhead.

Further, we do not take as input $\sigma_i$, but instead use the empirical variance of each arm.
\texttt{BMO UCB} relies heavily on the accuracy of its confidence intervals, which critically depend on the $\sigma_i$ used.
Too loose of a bound on the $\sigma_i$ increases running time undesirably, but too small of a $\sigma_i$ yields a large error probability.
For good theoretical performance, we require a good bound on the $\sigma_i$, but practically we estimate them by maintaining a (running) estimate of 
the mean and the second moment for every arm, and using the empirical variance as $\sigma_i^2$. 
Estimating confidence intervals using other techniques could potentially improve our algorithm practically, while determining ways to adaptively tighten bounds on the $\sigma_i$, like with empirical Bernstein, might tighten our analysis theoretically.

\subsection{Datasets}
For $k$-nearest neighbors and $k$-means 
we empirically evaluate the performance of \texttt{BMO-NN} on two real
world high-dimensional datasets: Tiny ImageNet \cite{ImageNet}, and $10x$ 
single cell RNA-Seq dataset \cite{10xdata}. 
Tiny ImageNet consists of $100$k images of size $64\times64$ with 
$3$ channels thus living in 
$12288$ dimensional space.
We downsize these images to evaluate our method on smaller dimensions.
We chose Tiny ImageNet because it is one of the most popular benchmark 
dataset in the field of computer vision -- where many applications use nearest
neighbors clustering in their pipelines. 
To further exhibit
the robustness of our approach across applications,
our second dataset is a single
cell RNA-Seq dataset from the field of computational
biology \cite{10xdata}. In addition to having both large dimensionality and sample size, it exhibits dramatic sparsity ($7\%$ nonzeros). For our empirical
evaluations we randomly sub-sample $100$k points from this dataset. We
note that each
point lies in $28$k dimensional space.

\subsection{Simulation Details}
We evaluate the accuracy of the algorithms
as follows:

\begin{enumerate}
    \item \textbf{$k$-NN: } For $n$ points, 
    let the true $k$-NN of point $i$ be the set $\texttt{NN}_i^{*}$
    and let the answer given by an algorithm be
    the set $\texttt{NN}_i^{\text{ALG}}$. We define the accuracy by 
    $\frac{1}{n}\sum_{i=0}^n\mathbb{E}\Big[ \mathbf{1}\big\{\texttt{NN}_i^{*}=\texttt{NN}_i^{\text{ALG}} \big\} \Big]$.
    $k=5$ was used for all simulations.

    \item \textbf{$k$-means}: For $n$ points and $k$ cluster centers,
    let the nearest cluster for point $i$ be $c_i^{*}$ and the returned answer be $c_i^{\text{ALG}}$. We define the accuracy by
    $\frac{1}{n}\sum_{i=0}^n\mathbb{E}\Big[\mathbf{1}\big\{ c_i^{*}=c_i^{\text{ALG}} \big\} \Big]$. 
    $k=100$ was used for all simulations.
\end{enumerate}
In the above expressions, the expectation is taken over the randomness in the sampling 
to obtain the estimators.
We evaluate the algorithms based on the number of coordinate-wise distance computations they make.

For our exact computation baseline we used scikit-learn's \texttt{NearestNeighbors} method.
Implementations for other algorithms were taken from the authors; NGT and kGraph were taken from the ANN benchmark of \cite{ann-benchmarks}, and Falconn’s from the FALCONN github \cite{andoni2015falconn_LSH}.
We tuned the ``number of probes" parameter in Falconn to achieve a desired accuracy of $99\%$.
We used the default parameters for NGT which yielded accuracies between $93\%$-$99\%$.
We adjusted iterations, K, and S parameters in kGraph to obtain the desired  accuracy of $99\%$.
For each algorithm, simulation points were generated by randomly selecting 1000 data points from the $n$ total, computing their nearest neighbors, and averaging the results.
For 1000 points, the confidence intervals were negligible, and so were not included.

\texttt{BMO-NN} was run with $\delta= .01$, initialized with 32 pulls per arm, and proceeded by pulling the top 32 arms simultaneously 256 times each per iteration (these were not optimized for specific operating points). 
Experiments were run on one core of an AMD Opteron Processor 6378 with 500GB memory (no parallelism).
Timing / FLOPS were only measured for querying, index construction was not included (note that \texttt{BMO-NN} does not require an index).
\texttt{BMO-NN} was run as described in Appendix \ref{sec:implementation}, achieving accuracy $>99\%$. For wall-clock comparisons we ignore the index construction time and only compare the querying time of other algorithms with \texttt{BMO-NN}'s querying time.
For number of coordinate-wise distance computations for LSH, we only consider the final exact computation step of Falconn.
In their implementation, hash tables are used to create a candidate set of nearest neighbors, and then the exact distance is computed to each of these points.
Hence, we lower bound the number of coordinate-wise distance computations LSH makes as $d\times$\texttt{size of candidate set}.
NGT and kGraph both output the exact number of distance computations made, which we multiply by $d$ to obtain the number of coordinate-wise distance computations made by the algorithms.

\end{document}